%% file: arxiv.tex
\documentclass{article} % For LaTeX2e
\usepackage{arxiv}
\usepackage{graphicx}
\usepackage{booktabs}
\usepackage[accsupp]{axessibility}  % Improves PDF readability for those with disabilities.
\usepackage[pagebackref,breaklinks]{hyperref}
\usepackage{orcidlink}
\usepackage{colortbl}
\definecolor{light-light-gray}{gray}{0.92} 
\usepackage{wrapfig}
\usepackage{algorithm}
\usepackage{algpseudocode}
\usepackage[normalem]{ulem}
\usepackage{array}
\usepackage{subcaption}

\input{math_commands.tex}

\usepackage{graphicx}
\usepackage{hyperref}
\usepackage{url}
\usepackage{booktabs}
\usepackage{multirow} 
\usepackage{booktabs}
\usepackage{makecell}
\usepackage[framemethod=default]{mdframed}
\usepackage[strict]{changepage}
\usepackage{enumitem}

\newcolumntype{R}[1]{>{\raggedleft\arraybackslash}p{#1}}

\newmdenv[
  linewidth=0pt,
  linecolor=black,
  innerleftmargin=5pt,
  innerrightmargin=5pt,
  skipabove=5pt,
  skipbelow=5pt
]{promptbox}

\newmdenv[
  linewidth=1pt,
  linecolor=black,
  topline=true,
  bottomline=true,
  leftline=true,
  rightline=true,
  innerleftmargin=10pt,
  innerrightmargin=10pt,
  innertopmargin=10pt,
  innerbottommargin=10pt,
  skipabove=1pt,
  skipbelow=1pt
]{examplebox}

\title{Architecture-Dependent Fusion Pathways in MLLMs}

\author{ {\hspace{0.1mm}Hebao Zhu}\\
    \texttt{Mohamed bin Zayed University of Artificial Intelligence}\\
    \texttt{Abu Dhabi, UAE}\\
	\texttt{hebao.zhu@mbzuai.ac.ae} \\
    \And{\hspace{0.1mm}Dongxia Wu}\\
    \texttt{Mohamed bin Zayed University of Artificial Intelligence}\\
    \texttt{Abu Dhabi, UAE}\\
	\texttt{dongxia.wu@mbzuai.ac.ae} \\
}

\date{}

\begin{document}

\maketitle

% \blfootnote{$^*$ Equal contribution. $^\dagger$ Equal advising.}

% \input{secs/0_abs}

% \input{secs/1_intro}
% \input{secs/5_related}
% \input{secs/2_problem}
% \input{secs/3_method}
% \input{secs/4_exp_arxiv}

\begin{abstract}
Multimodal Large Language Models (MLLMs) achieve strong performance across vision-language tasks, yet the internal mechanisms by which visual and textual information are fused across layers remain insufficiently understood. 
We investigate representative MLLMs from two architectural paradigms: concatenation architectures and native multimodal architectures. 
We conduct three progressively connected analyses: alignment decoupling identifies which modality changes, attention routing and entropy characterize how cross-modal information is distributed, and intrinsic dimensionality examines how fusion reshapes feature spaces. 
Separately, we perform causal intervention experiments as a validation of the resulting interpretation. 
As a supplementary analysis, we use visual CKA to examine the Platonic Representation Hypothesis. 
Together, these analyses reveal two distinct fusion pathways: concatenation models follow a text-first, vision-later pathway, whereas native models exhibit earlier visual-textual co-adaptation and feature-space reorganization. 
This work provides a mechanistic perspective for understanding multimodal fusion and supports architecture-aware diagnostics of multimodal representations.
\end{abstract}

\section{Introduction}

MLLMs have rapidly advanced a wide range of vision-language applications, including human-computer interaction~\cite{achiam2023gpt}, medical image analysis~\cite{tu2024towards}, and autonomous driving~\cite{sima2024drivelm}.
Despite strong empirical progress, most studies still evaluate MLLMs primarily through downstream task accuracy and multimodal reasoning benchmarks~\cite{zhu2023minigpt,wang2024qwen2}.
Final performance, however, does not reveal how visual and textual tokens interact inside the model.
The question is therefore not only whether an MLLM solves a task, but also how it fuses modalities across depth.

A central factor behind this problem is architectural diversity.
Existing MLLMs can be broadly categorized into two paradigms according to how visual and textual information are integrated~\cite{yin2024survey}: \emph{concatenation architectures} and \emph{native multimodal architectures}.
We refer to Section~\ref{sec:architectural_mllms} for the architectural definitions and representative examples.
Although these paradigms differ in how they introduce and process visual information, their layer-wise fusion dynamics have not been systematically characterized.

% Recent studies suggest that understanding MLLMs requires moving beyond final accuracy and examining representation alignment and cross-modal interaction.
Prior work has revealed modality gaps in multimodal representation learning~\cite{liang2022mind}, shown that attention mechanisms shape multimodal representations~\cite{hendricks2021decoupling}, and demonstrated that neural representations evolve substantially across layers~\cite{phang2021fine,lee2025multimodal}.
Motivated by these findings, we first conduct a layer-wise cross-modal representation analysis using Centered Kernel Alignment (CKA)~\cite{kornblith2019similarity}.
CKA measures the similarity between two feature matrices and is widely used for comparison in multimodal analysis because it is robust to orthogonal transformations and isotropic scaling.

\begin{figure}[t]
    \centering
    \includegraphics[width=\linewidth]{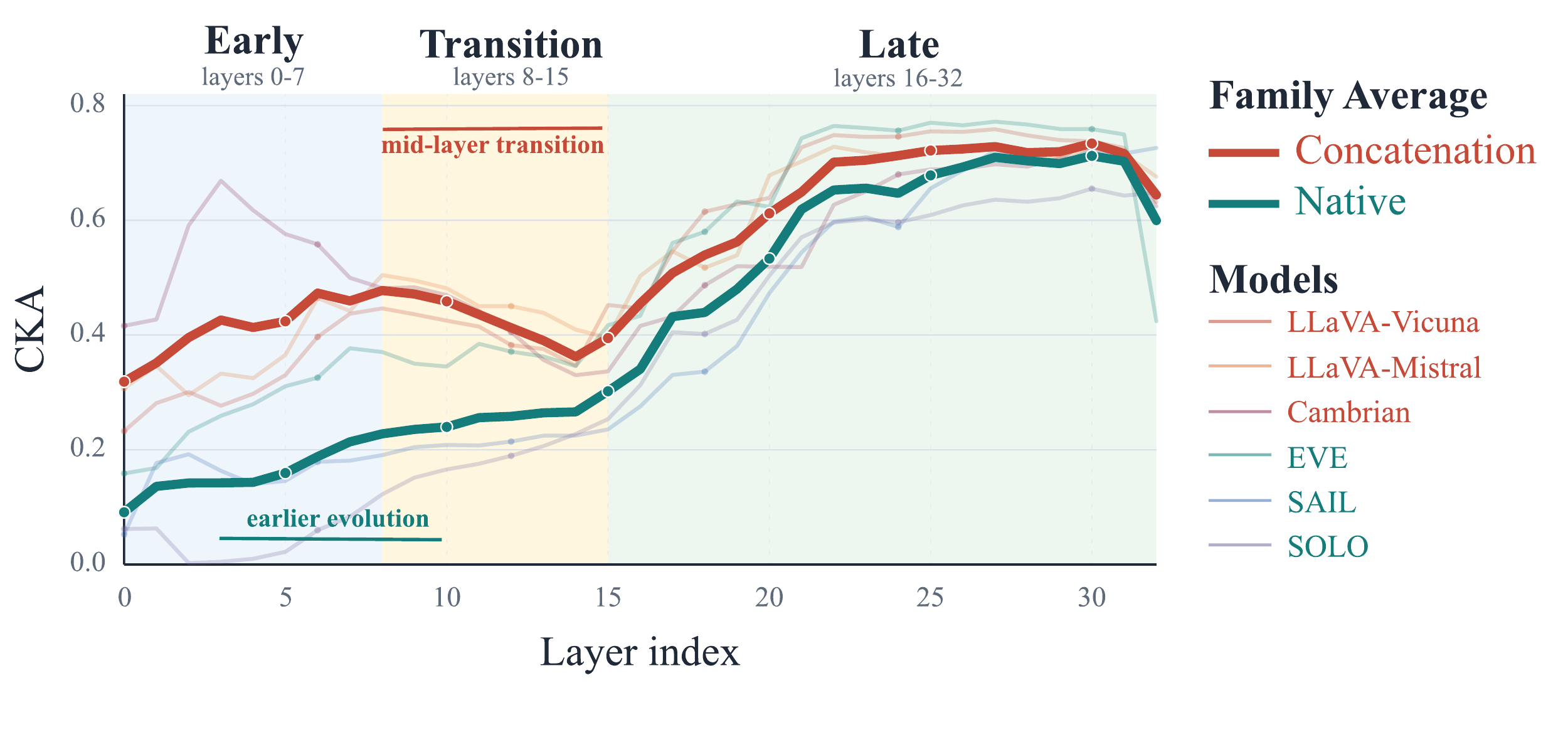}
    % \vspace{-6mm}
    \caption{Cross-modal CKA similarity across layers. \textbf{{Family average}} represents the average values of different types of models. Shaded regions mark the early, transition, and late layer ranges. The model classification is described in EXPERIMENTAL SETUP~\ref{sec:experiment_setup}. 
    Model-name colors distinguish concatenation from native architectures.}
    \label{fig:cka_comparison}
\end{figure}

As illustrated in Figure~\ref{fig:cka_comparison}, 
concatenation architectures generally exhibit relatively high similarity in early layers, although the magnitude depends on the employed vision encoder.
These models then experience a pronounced decline in intermediate layers, suggesting a temporary divergence between modalities during feature transformation.
In contrast, native multimodal architectures begin changing earlier and maintain cross-modal interaction through the early and intermediate stages.
Beyond the transition range, similarity generally recovers in middle-to-deep layers, with model-specific declines near the final layers.
This phenomenon suggests that different architectures follow distinct fusion pathways before eventually forming more similar high-level representations.

Cross-modal CKA alone is insufficient to explain the mechanism behind this pattern.
First, it is an aggregate similarity score and cannot determine which modality is primarily changing during fusion.
Second, it does not reveal how information is routed between visual and textual tokens.
Third, it cannot distinguish whether representation changes correspond to feature compression, expansion, or structural reorganization.
Finally, representation-level analysis alone does not establish whether the identified transition layers are functionally important.

We therefore organize the study around three progressively connected analyses.
First, alignment decoupling separately tracks the layer-wise evolution of visual and textual representations.
Second, attention routing and entropy reveal how information flows between modalities and whether cross-modal attention is concentrated or diffuse.
Third, intrinsic dimensionality measures whether fusion corresponds to feature-space compression, expansion, or reorganization.
Separately, causal intervention tests whether the identified routing and layer-transition patterns are functionally relevant.
Together, the three analyses reveal two fusion pathways: concatenation models follow a text-first, vision-later pathway, whereas native models show earlier visual-textual co-adaptation.

Our main contributions are summarized as follows:
\begin{itemize}
    \item \textbf{Three progressive experiments.} We conduct three related experiments that move from which modality changes, to how cross-modal information is routed, to how feature spaces reorganize, revealing two distinct fusion pathways.
    \item \textbf{Causal validation.} We separately conduct intervention experiments to test whether the mechanisms identified above are functionally relevant.
    \item \textbf{Supplementary Platonic analysis.} We conduct a visual CKA experiment to test whether deep visual representations across MLLMs become more aligned, providing evidence relevant to the Platonic Representation Hypothesis~\cite{huh2024platonic}.
\end{itemize}

\section{PRELIMINARY}

\begin{figure}[t]
    \centering
    \includegraphics[width=\linewidth]{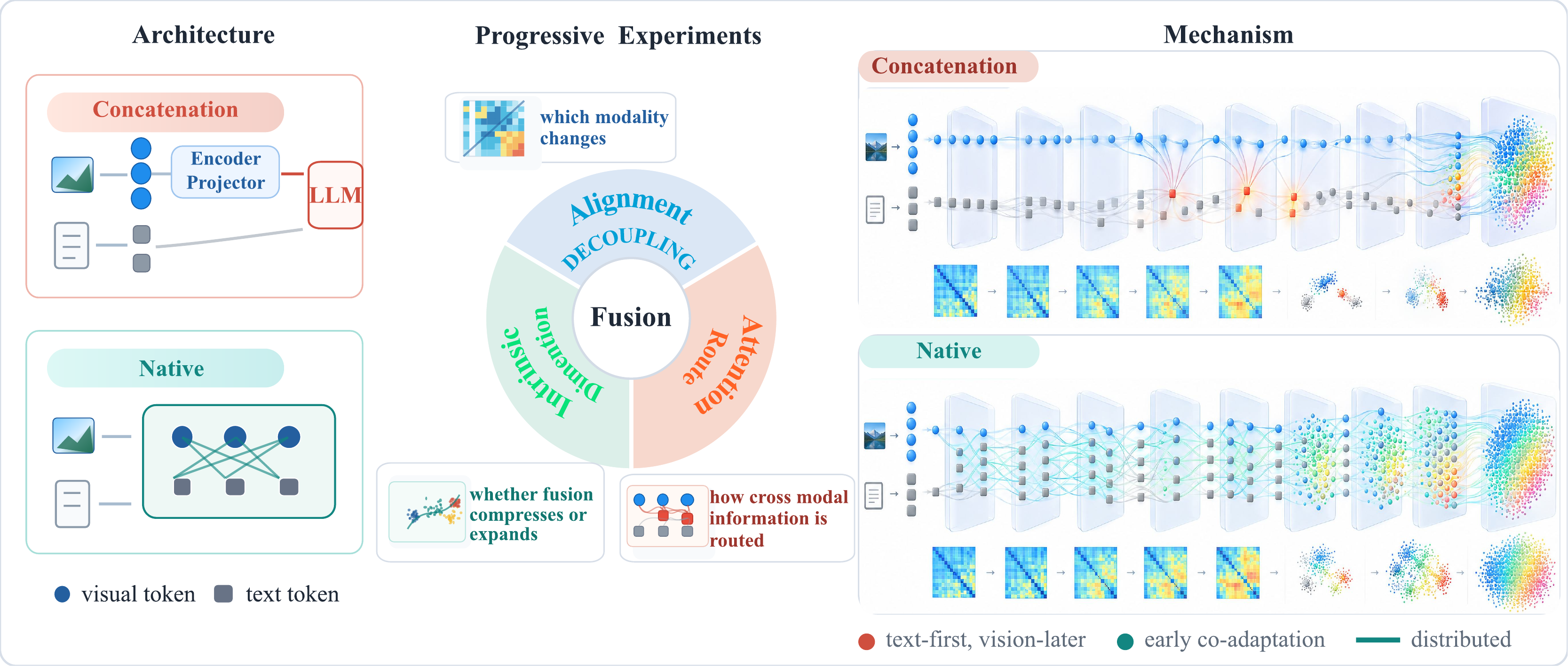}
    % \vspace{-5mm}
    \caption{Schematic overview of the study design. The left side contrasts concatenation and native multimodal architectures, the center groups the three progressive experiments, and the right side summarizes the two fusion pathways suggested by these analyses.}
    \label{fig:fusion_summary}
\end{figure}
% \vspace{-6pt}

\subsection{MLLM Architectures}
\label{sec:architectural_mllms}

MLLMs can generally be divided into concatenation architectures and native multimodal architectures according to the design of the vision-language interface~\cite{yin2024survey}.
Concatenation architectures preserve the original language model backbone while introducing lightweight projection modules to align visual features with the textual embedding space.
Most of these frameworks rely on pretrained CLIP visual encoders~\cite{radford2021learning}, whose large-scale contrastive pretraining provides transferable semantic representations for multimodal alignment.
The core assumption of such architectures is that the visual encoder already provides sufficiently high-level semantics for visual tokens~\cite{li2023blip}.

Native multimodal architectures learn visual tokens and language representations within a shared end-to-end multimodal framework.
Their visual front ends may be jointly optimized or replaced by encoder-free visual tokenization, so the category is defined by the multimodal learning interface rather than by a single tokenization implementation.
Our work is complementary: rather than treating the empirical difference between architectures as self-evident, we examine the internal mechanisms that give rise to different fusion pathways.

\subsection{MLLM Alignment and Interpretability}
\label{sec:alignment}

Recent works often adopt the heuristic that stronger alignment leads to better multimodal performance~\cite{li2021align,almudevar2025aligning,singh2022flava}.
However, recent evidence suggests that stronger alignment does not always lead to better performance~\cite{fang2026align}.
Other studies analyze internal information flow in MLLMs but remain focused primarily on concatenation architectures~\cite{zhang2025cross} or on representation alignment alone~\cite{lu2025representation}.
Existing probing-based analyses often focus on isolated layer representations rather than the complete evolution of multimodal fusion~\cite{phang2021fine}.
In summary, prior work has not yet provided a comprehensive explanation of how different MLLM architectures fuse modalities across depth.

\section{Experiments}
\label{sec:experiments}

\subsection{Experimental Setup}
\label{sec:experiment_setup}

To explain the cross-modal CKA phenomenon, we design three progressively connected analyses from the perspectives of representation alignment, attention routing and entropy, and intrinsic feature-space structure.
The first identifies \emph{which modality changes}, the second explains \emph{how the change is mediated}, and the third examines \emph{whether the change corresponds to compression or expansion}.
We then conduct a separate causal intervention validation to test \emph{whether the identified mechanism is functionally relevant}.

\paragraph{Model Selection.}
We evaluate three concatenation models, LLaVA-next with Vicuna and Mistral backbones~\cite{liu2024llavanext} and Cambrian~\cite{tong2024cambrian}, and three native multimodal models, SOLO~\cite{chen2024solo}, EVE~\cite{diao2024unveiling}, and SAIL~\cite{lei2025scalability}.
The LLaVA series uses CLIP-ViT-L~\cite{radford2021learning} as its visual encoder, whereas Cambrian is a distinct concatenation model employing a hybrid visual encoder that integrates DINO~\cite{oquab2023dinov2} and CLIP~\cite{radford2021learning}.
Because the two architectural paradigms use fundamentally different training pipelines, a perfectly controlled comparison that fixes every factor is not feasible.
Concatenation models typically combine a pretrained visual encoder, a projection module, and a language model backbone, while native models learn visual tokenization and language representations in an end-to-end multimodal framework with model-specific supervision pipelines.
Our goal is to identify paradigm-level regularities under representative open models.
To reduce single-model artifacts, we evaluate multiple models per paradigm and require the same interpretation to be supported across CKA, attention, intrinsic-dimensionality, and intervention evidence.

\paragraph{Dataset and Input Construction.}
We randomly sample 10,000 image-text pairs from COCO-val~\cite{lin2014microsoft}, VQA~\cite{antol2015vqa}, and MMMU~\cite{yue2024mmmu} to cover different levels of visual complexity, textual semantic density, and multimodal reasoning requirements.
For the representation and attention analyses, we use clean caption-style image-text inputs to reduce noise from long reasoning chains or text spans that are only weakly grounded in the image.
All special tokens, padding tokens, and truncation positions are recorded and masked during analysis to ensure that the measured statistics reflect only effective visual and textual tokens.
We evaluate complex reasoning behavior separately through the causal perturbation experiment on MMBench~\cite{liu2024mmbench} using VLMEvalKit~\cite{duan2024vlmevalkit}.

\paragraph{Feature and Attention Extraction.}
\label{par:extraction}
For each model and each input sample, we extract the post-norm hidden representation matrix
$H^{(l)} \in \mathbb{R}^{N_l \times d}$ from every layer, where $N_l$ denotes the number of valid tokens and $d$ denotes the hidden dimension.
In all experiments, the analyzed hidden representations are normalized layer by layer and use a common hidden size of $d=4096$.
The numbers of visual and textual tokens follow the default configuration of each model, so the analysis preserves the native tokenization and visual-token construction of different architectures.
According to the token indices of each model, we separate the hidden states into visual and textual token sets, denoted as $H_V^{(l)}$ and $H_T^{(l)}$, respectively.
Layer indices are zero-based (0--32), with index 0 denoting the initial CKA reference representation.
We also extract post-softmax attention matrices from all layers and heads.

\paragraph{Statistical Robustness.}
To quantify sampling uncertainty, we conduct non-parametric bootstrap resampling of pairs on a 5,000-pair subset drawn from population used in the main representation analysis.
For each sample-size group, we perform 1,000 independent resamplings with replacement, compute the target value for each layer, and report bootstrap mean, standard error, and relative bias.
Appendix~\ref{app:bootstrap} reports the detailed results for Cambrian at layers 1, 15, and 30.

\subsection{Alignment Decoupling}
\label{sec:alignment_decoupling}

\paragraph{Motivation.}
The global cross-modal CKA curve in Figure~\ref{fig:cka_comparison} shows that different architectures follow distinct fusion trajectories.
However, cross-modal CKA alone is an aggregate measure and cannot determine which modality is primarily responsible for the observed change.
A decrease in visual-textual similarity may indicate that visual tokens are being reshaped by the language manifold, that textual tokens are adapting to visual semantics, or that both modalities are changing simultaneously.
This analysis answers the first mechanistic question: \emph{during multimodal fusion, who is moving toward whom?}

\paragraph{Method.}
In addition to the cross-modal similarity $\operatorname{CKA}_{V,T}^{(l)}$, we compute intra-modality CKA relative to the initial representation (index 0).
For each input sample, visual and textual token representations are first averaged within their corresponding modality to obtain sample-level pooled representations, which makes representations comparable when different models or modalities have different token numbers:
\[
\operatorname{CKA}_{V,V}^{(l)}
= \operatorname{CKA}\left(H_V^{(0)}, H_V^{(l)}\right),
\qquad
\operatorname{CKA}_{T,T}^{(l)}
= \operatorname{CKA}\left(H_T^{(0)}, H_T^{(l)}\right).
\]
Here, $\operatorname{CKA}_{V,V}^{(l)}$ measures how much the visual representation changes, while $\operatorname{CKA}_{T,T}^{(l)}$ measures the corresponding evolution of textual representations.
A higher value indicates that the modality retains a representation close to its initial state, whereas a lower value indicates stronger layer-wise deformation.

\begin{figure}[t]
    \centering
    \includegraphics[width=\linewidth]{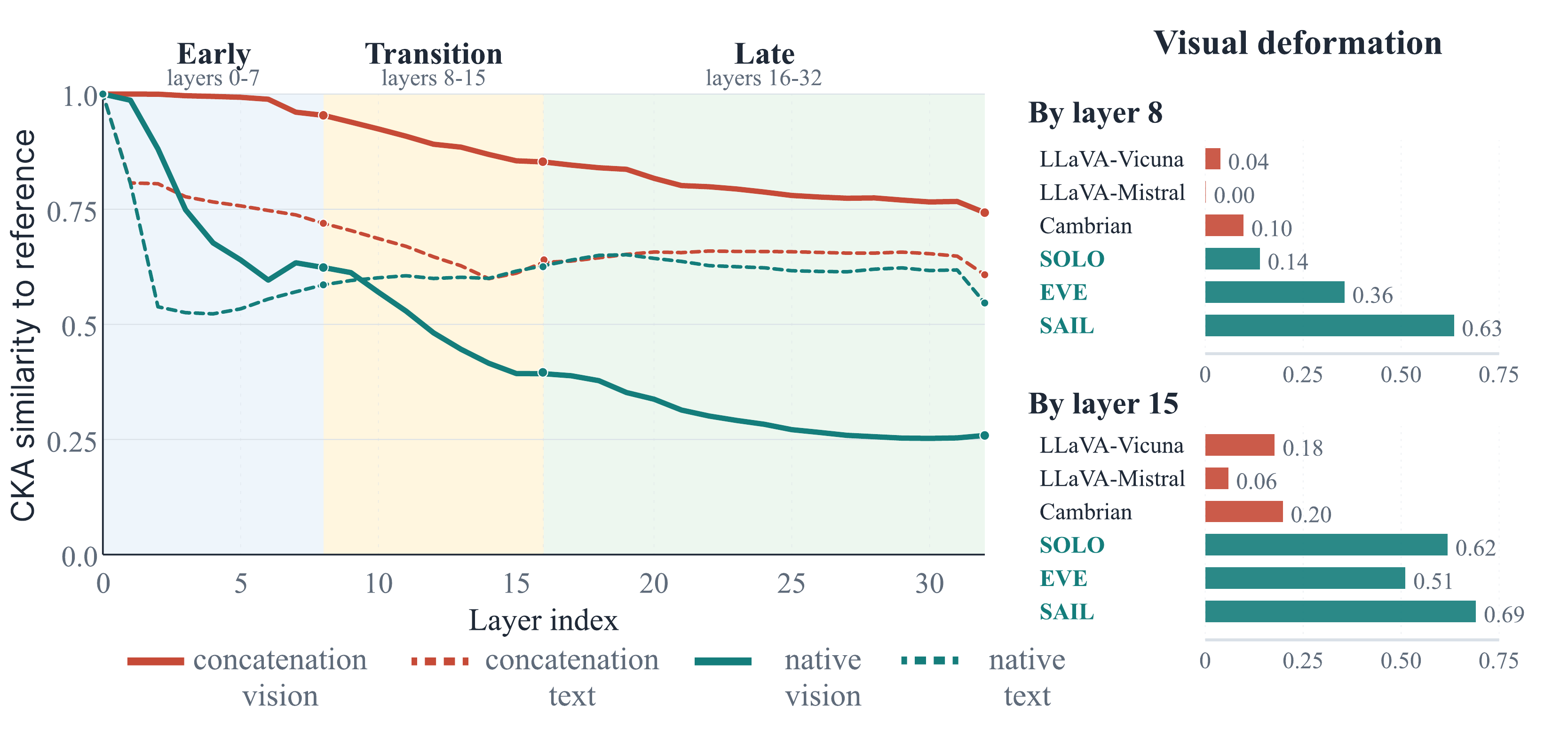}
    % \vspace{-5mm}
    \caption{Intra-modality CKA relative to the initial reference representation. Left: Solid and dashed lines show family-average visual and textual CKA, respectively. Right: Per-model visual deformation ($1-\operatorname{CKA}_{V,V}$), at displayed layer indices 8 and 15. Individual-model curves are provided in Appendix~\ref{app:intra_modality_cka}.}
    \label{fig:intra_modality_cka}
\end{figure}

\paragraph{Results.}
Figure~\ref{fig:intra_modality_cka} illustrates distinct differences between concatenation and native multimodal architectures.
For concatenation models, visual curves stay close to the initial reference representation in shallow layers, meaning visual tokens largely retain the structure from the pretrained visual encoder at the input of the language model stack. 
By contrast, textual representations deform earlier and stabilize in middle layers. Following this textual adaptation, visual curves drop more evidently, indicating late reshaping of visual representations. 
The per model bars at layers 8 and 15 further show that concatenation models maintain higher visual CKA relative to native models.

Native multimodal architectures follow a different trend: their visual representations diverge from the initial reference state much earlier, while textual representations evolve with depth with model-specific curve patterns. 
Thus, the core finding is not perfectly synchronized cross-modality transformation across all layers, but that native models trigger earlier and more continuous visual-textual co-adaptation compared with concatenation models.
The individual visual and textual trajectories in Figure~\ref{fig:app_intra_modality_cka} further show the model-specific patterns underlying these family averages.

\subsection{Attention Routing and Entropy}
\label{sec:attention_routing_entropy}

\paragraph{Motivation.}
The previous experiment identifies which modality changes during fusion, but it does not explain how such changes are implemented inside the model.
Since MLLMs are transformer-based architectures, attention is the primary operator through which tokens exchange information.
Routing ratios measure how much attention mass one modality allocates to the other~\cite{abnar2020quantifying}, while entropy measures whether this attention is concentrated on a small number of tokens or dispersed across many tokens~\cite{clark2019does}.
This analysis answers the second mechanistic question: \emph{through what information-flow pattern does cross-modal fusion occur?}

\begin{figure}[t]
    \centering
    \includegraphics[width=\linewidth]{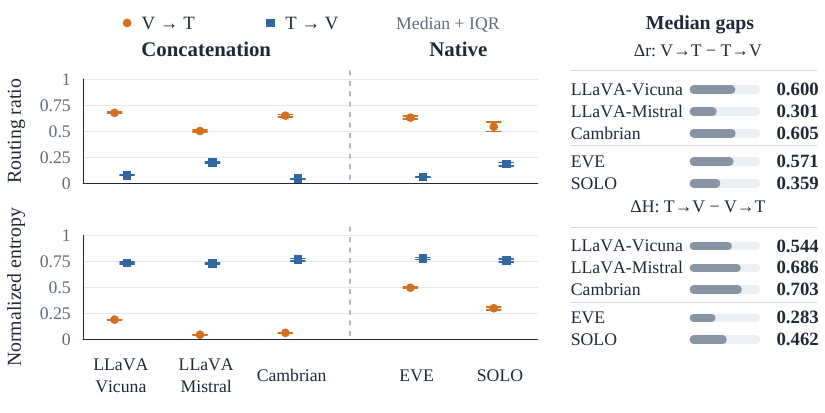}
    % \vspace{-5mm}
    \caption{Cross-modal attention over layers 8--15. Top: routing ratio; bottom: normalized entropy. Orange circles denote $V\rightarrow T$ and blue squares denote $T\rightarrow V$, where arrows indicate query-to-key attention allocation. 
    Left: medians and interquartile ranges of per-pair layer averages. Right: differences between the displayed medians, $\Delta r$ ($V\rightarrow T$ minus $T\rightarrow V$) and $\Delta H$ (the reverse subtraction). 
    SAIL is omitted because its hybrid attention path constrains visual-to-text entropy. 
    Layer-wise curves for all six models appear in Appendix~\ref{app:attention_routing_entropy}.}
    \label{fig:attention_routing_entropy}
\end{figure}

\paragraph{Method.}
Given visual token set $V$ and textual token set $T$, we define the routing mass from query modality $M$ to key modality $N$ as
\[
r^{(l)}_{M\rightarrow N}
=
\frac{1}{|M|}
\sum_{i\in M}\sum_{j\in N}\bar{A}^{(l)}_{ij},
\qquad
M,N\in\{V,T\},\; M\neq N.
\]
Here, $r^{(l)}_{V\rightarrow T}$ measures how much attention mass visual query tokens allocate to textual keys, while $r^{(l)}_{T\rightarrow V}$ measures the reverse direction.
\mbox{We normalize over target tokens:}
\[
\tilde{A}^{(l,M\rightarrow N)}_{ij}
=
\frac{\bar{A}^{(l)}_{ij}}
{\sum_{j'\in N}\bar{A}^{(l)}_{ij'}+\epsilon},
\qquad j\in N.
\]
Since different modalities may contain different numbers of target tokens, we further use the normalized cross-modal entropy
\[
\widehat{H}^{(l)}_{M\rightarrow N}
=
-\frac{1}{|M|}
\sum_{i\in M}
\frac{
\sum_{j\in N}
\tilde{A}^{(l,M\rightarrow N)}_{ij}
\log\left(\tilde{A}^{(l,M\rightarrow N)}_{ij}+\epsilon\right)
}
{\log(|N|+\epsilon)}.
\]
A lower $\widehat{H}^{(l)}_{M\rightarrow N}$ indicates more concentrated cross-modal attention, while a higher value indicates more diffuse attention over the target modality.
In the implementation, entropy is normalized for each sample using its actual target-token count before aggregation.
For Figure~\ref{fig:attention_routing_entropy}, we first average each pair's routing ratio and normalized entropy across layers 8--15, then report the median and interquartile range across pairs.
The right-hand summaries report differences between the displayed directional medians: visual-to-text minus text-to-vision for routing ($\Delta r$), and text-to-vision minus visual-to-text for entropy ($\Delta H$).

\paragraph{Results.}
Figure~\ref{fig:attention_routing_entropy} shows directional routing asymmetry in all displayed models: the median visual to text routing ratio exceeds the reverse-direction median, with $\Delta r$ ranging from 0.301 to 0.605.
Large routing gaps occur in both architectural families; for example, Cambrian and EVE have gaps of 0.605 and 0.571, respectively.
Routing magnitude alone therefore does not clearly separate the two families.

Attention concentration provides a more differentiated comparison.
The concatenation models have median visual-to-text normalized entropies of 0.046--0.192, compared with 0.497 for EVE and 0.301 for SOLO.
Their directional entropy gaps are also larger: $\Delta H=0.544$--$0.703$ for the concatenation models, versus 0.283 for EVE and 0.462 for SOLO.
These observations are consistent with more concentrated visual-query attention to textual keys in the concatenation models, and more diffuse attention in this direction in EVE and SOLO.

\subsection{Intrinsic Dimensionality Transition}
\label{sec:intrinsic_dimension}

\paragraph{Motivation.}
Both CKA and attention statistics are relative measures.
CKA describes similarity between representations, and attention describes how tokens interact, but neither directly indicates whether the feature space is being compressed, expanded, or reorganized during fusion.
For example, a decrease in CKA may correspond either to information loss, feature specialization, or expansion into a richer semantic subspace.
To address this limitation, we analyze the intrinsic dimensionality of visual and textual representations.
This provides an absolute measure of the effective complexity of the feature space and answers the third mechanistic question: \emph{what kind of structural transformation occurs during multimodal fusion?}

\paragraph{Method.}
Because the hidden representations are normalized layer by layer and compared at a fixed hidden size, the intrinsic-dimensionality curves are not caused by differences in hidden size or activation scale across layers.
The token numbers are kept as each model's default setting, while the same normalization and variance-threshold protocol is applied to all models and modalities.
For each modality $M \in \{V,T\}$ and each layer $l$, we perform SVD on the centered feature matrix.
Let $\sigma_i$ denote the $i$-th singular value and let $\tau$ denote the variance preservation threshold.
The intrinsic dimensionality is defined as
\[
\operatorname{ID}_{M,\tau}^{(l)}
=
\min
\left\{
k:
\frac{\sum_{i=1}^{k}\sigma_i^2}
{\sum_i \sigma_i^2}
\ge \tau
\right\}.
\]
We evaluate multiple thresholds $\tau \in \{0.5, 0.6, 0.7, 0.8\}$ to ensure that the observed trends are not caused by a particular variance cutoff.
A higher intrinsic dimensionality indicates that the token representations occupy a richer and more diverse effective subspace, while a lower value suggests a more compressed or redundant representation structure.

\begin{figure}[t]
    \centering
    \includegraphics[width=\linewidth]{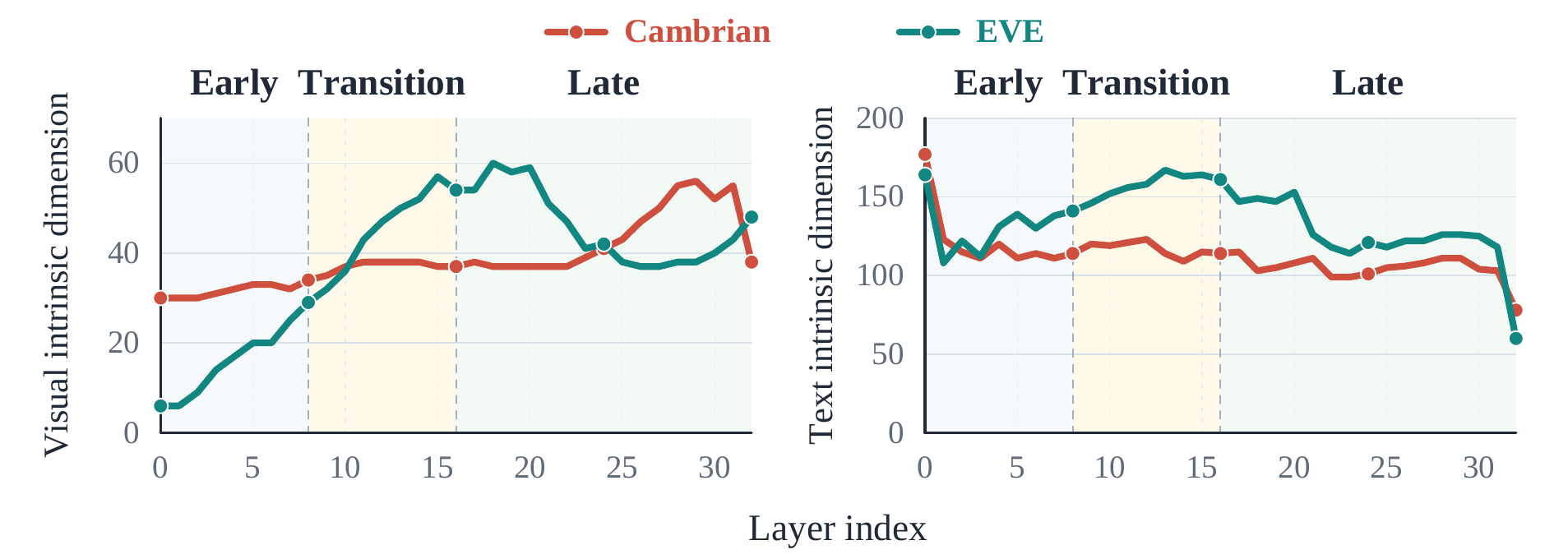}
    \caption{Intrinsic dimensionality at $\tau=0.7$ for representative concatenation (Cambrian) and native (EVE) models. The left and right panels report visual and textual intrinsic dimensionality on separate scales. Results for additional models and threshold sensitivity are reported in Appendix~\ref{app:intrinsic_dimension}.}
    \label{fig:intrinsic_dimension}
\end{figure}

\paragraph{Results.}
Figure~\ref{fig:intrinsic_dimension} shows that intrinsic dimensionality provides a structural view of the fusion process.
For the concatenation model Cambrian, the visual intrinsic dimension remains relatively low and stable through the early and transition ranges.
This plateau suggests that the visual tokens initially preserve the compact structure inherited from the pretrained visual encoder.
Around the middle-to-deep transition, the visual intrinsic dimension increases, indicating that visual representations expand into a richer effective subspace.
The textual intrinsic dimension decreases from its input-side value and then changes more gradually, suggesting early reorganization of the language stream followed by stabilization.

For the native model EVE, the visual intrinsic dimension begins to increase from earlier layers and does not exhibit the long shallow-layer plateau observed in concatenation models.
The textual intrinsic dimension changes together with the visual side and can peak around the transition before decreasing in later layers.
Therefore, native models are better characterized as performing early feature-space expansion and reorganization, rather than strictly monotonic expansion across all layers.

\subsection{Causal Intervention Validation}
\label{sec:causal_validation}

The preceding analyses identify recurring internal patterns, but intervention experiments are needed to test whether those patterns are functionally relevant.
We therefore conduct two complementary causal validations.
The first masks sparse textual anchors selected from attention-concentrated text tokens and evaluates object hallucination behavior on POPE~\cite{li2023evaluating}.
The second injects Gaussian noise into early-layer or late-layer hidden states and evaluates general multimodal question answering on MMBench using VLMEvalKit~\cite{liu2024mmbench,duan2024vlmevalkit}.
Early/Late denote layers 0--7/16--32 in Figure~\ref{fig:cka_comparison}; POPE scores are F1 (\%), and all drops are in percentage points.

\begin{table}[H]
\centering
\captionsetup{justification=raggedright,singlelinecheck=false}
\begin{minipage}[t]{0.40\linewidth}
\captionof{table}{Textual-target masking on POPE.}
\label{tab:anchor_masking}
\end{minipage}
\hfill
\begin{minipage}[t]{0.58\linewidth}
\captionof{table}{Layer perturbation on MMBench General VQA.}
\label{tab:layer_perturbation}
\end{minipage}
\par
\begin{minipage}[t]{0.40\linewidth}
\vspace{0pt}
\centering
\small
\renewcommand{\arraystretch}{1.10}
\setlength{\tabcolsep}{2.3pt}
\begin{tabular*}{\linewidth}{@{}l@{\extracolsep{\fill}}ccc@{}}
\toprule
Model & Original & Masked & Drop \\
\midrule
LLaVA-Vicuna & 87.5 & 44.6 & 42.9$\Downarrow$ \\
LLaVA-Mistral & 86.4 & 39.9 & 46.5$\Downarrow$ \\
Cambrian & 86.4 & 36.3 & 50.1$\Downarrow$ \\
EVE & 85.0 & 53.8 & 31.2$\Downarrow$ \\
SOLO & 78.6 & 43.8 & 34.8$\Downarrow$ \\
\bottomrule
\end{tabular*}
\par\vspace{3pt}
\raggedright
SAIL is omitted due to its hybrid attention mechanism.
\end{minipage}
\hfill
\begin{minipage}[t]{0.58\linewidth}
\vspace{0pt}
\centering
\small
\renewcommand{\arraystretch}{1.10}
\setlength{\tabcolsep}{2.8pt}
\begin{tabular*}{\linewidth}{@{}l@{\extracolsep{\fill}}ccccc@{}}
\toprule
Model & Orig. & Early & E.drop & Late & L.drop \\
\midrule
LLaVA-Vicuna & 66.5 & 53.1 & 13.4$\Downarrow$ & 46.7 & 19.8$\Downarrow$ \\
LLaVA-Mistral & 66.5 & 53.4 & 13.1$\Downarrow$ & 47.0 & 19.5$\Downarrow$ \\
Cambrian & 75.9 & 67.0 & 8.9$\Downarrow$ & 58.0 & 17.9$\Downarrow$ \\
SAIL & 70.1 & 49.5 & 20.6$\Downarrow$ & 58.6 & 11.5$\Downarrow$ \\
EVE & 52.3 & 29.9 & 22.4$\Downarrow$ & 38.6 & 13.7$\Downarrow$ \\
SOLO & 67.7 & 43.6 & 24.1$\Downarrow$ & 56.8 & 10.9$\Downarrow$ \\
\bottomrule
\end{tabular*}
\end{minipage}
\end{table}

Table~\ref{tab:anchor_masking} shows that concatenation models suffer larger POPE drops when concentrated textual targets are masked.
This supports the attention-routing interpretation that their visual stream relies on a small set of textual targets during the transition stage.
Table~\ref{tab:layer_perturbation} shows the complementary MMBench pattern: native models are more vulnerable to early-layer noise, whereas concatenation models are more vulnerable to late-layer noise.
This is consistent with the proposed pathways: native architectures begin visual-textual co-adaptation early, while concatenation architectures defer substantial visual reshaping to later transition layers.

\subsection{Mechanistic Summary}
\label{sec:analysis}

The three progressive analyses provide an explanation for the fusion.
Alignment decoupling shows that concatenation architectures follow a sequential pathway: textual representations adapt earlier, while visual representations remain relatively stable in shallow layers and are reshaped later.
Within layers 8--15, attention routing shows that concatenation models allocate substantial visual-query attention to textual keys, with lower normalized entropy in this direction than in the reverse direction.
Intrinsic dimensionality reveals the structural nature of this process: the later deformation of visual representations corresponds to delayed feature-space expansion.
Therefore, concatenation models can be summarized as following a text-first, vision-later fusion pathway.

Native architectures exhibit a different pattern.
Their visual representations change from earlier layers, and textual representations also evolve with depth.
The attention summaries show model-specific directional asymmetries, including more diffuse visual-to-text attention in EVE and SOLO than in the concatenation models, with SAIL forming a structural exception documented in Appendix~\ref{app:attention_routing_entropy}.
Their intrinsic dimensionality curves suggest earlier feature-space expansion and reorganization, and their stronger sensitivity to early perturbation supports the view that fusion begins earlier.
Thus, the native-model results support earlier visual-textual co-adaptation while also revealing heterogeneous attention profiles.
Separately, the causal intervention results support the functional relevance of these two pathway interpretations.

\section{Deep Visual Representation Convergence}
\label{sec:visual_representation_convergence}

To further examine whether different MLLMs converge toward similar high level representations, we conduct a supplementary analysis from the perspective of representation similarity.
This experiment is motivated by the Platonic Representation Hypothesis~\cite{huh2024platonic}, which suggests that representations learned by different neural networks may gradually converge toward structurally similar latent spaces.
Here, we use layer-wise visual CKA similarity to test whether such a convergence trend can be observed across MLLMs with different architectures.

\paragraph{Method.}
We compute pairwise CKA similarity between the visual representation spaces of different models:
\[
\mathcal{S}_{ij}^{(l)}
=
\operatorname{CKA}
\left(
Z_i^{(l)}, Z_j^{(l)}
\right),
\]
where $Z_i^{(l)}$ and $Z_j^{(l)}$ denote the stacked visual representations of two models at layer $l$.
This produces a model-by-model visual CKA matrix for each layer.

\begin{figure}[t]
    \centering
    \begin{subfigure}[t]{0.32\linewidth}
        \centering
        \includegraphics[width=\linewidth]{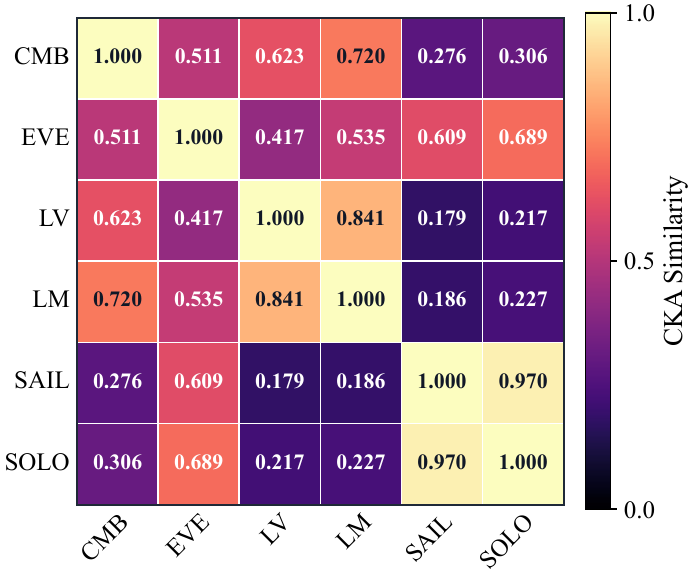}
        \caption{Layer 1.}
        \label{fig:vision_cka_layer1}
    \end{subfigure}
    \hfill
    \begin{subfigure}[t]{0.32\linewidth}
        \centering
        \includegraphics[width=\linewidth]{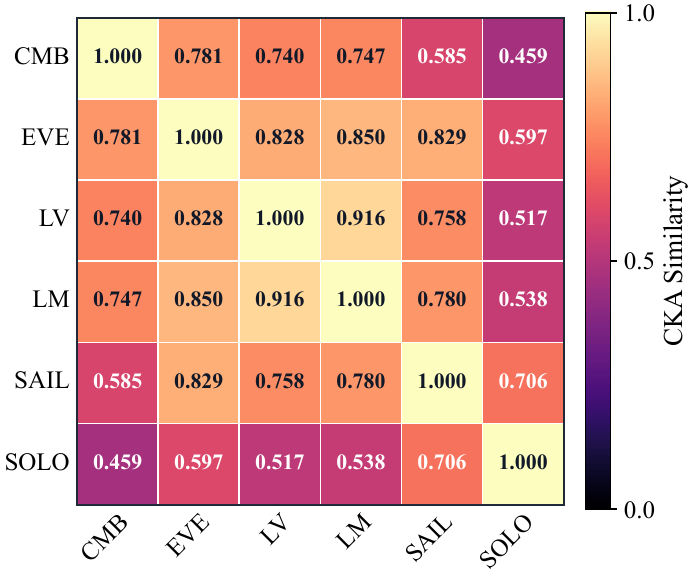}
        \caption{Layer 15.}
        \label{fig:vision_cka_layer15}
    \end{subfigure}
    \hfill
    \begin{subfigure}[t]{0.32\linewidth}
        \centering
        \includegraphics[width=\linewidth]{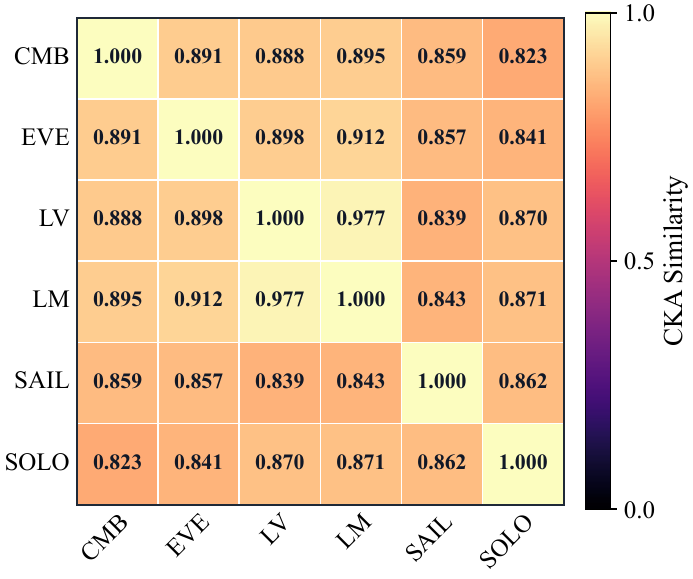}
        \caption{Layer 30.}
        \label{fig:vision_cka_layer30}
    \end{subfigure}
    \caption{Cross-model visual CKA heatmaps at shallow, transition, and deep layers. Each cell measures the similarity between the visual representation spaces of two MLLMs at the same layer. CMB, LV, and LM denote Cambrian, LLaVA-Vicuna, and LLaVA-Mistral. Additional selected layers are reported in Appendix~\ref{app:cross_model_visual_similarity}.}
    \label{fig:cross_model_vision_cka}
\end{figure}

\paragraph{Results.}
As shown in Figure~\ref{fig:cross_model_vision_cka}, visual representations are highly architecture-dependent at layer 1.
Models with similar visual processing strategies show relatively high similarity, while cross-paradigm similarities remain low.
This indicates that shallow visual representations are mainly determined by each model's visual encoder, tokenizer, and architectural design.
In contrast, the deep-layer heatmap at layer 30 shows substantially higher similarity across almost all model pairs.
This suggests that although different MLLMs start from different visual representation spaces, their deep visual representations gradually move toward a more similar semantic structure.

These results provide empirical evidence consistent with the Platonic Representation Hypothesis in the context of MLLMs.
From the perspective of representation similarity, multimodal fusion appears to reduce architecture-specific differences and encourages different models to form more aligned high-level visual representations.

\section{Conclusion}
\label{sec:conclusion}

The main contribution of this work is to uncover internal fusion mechanisms and design experiments to validate them. The findings can suggest several concrete directions.
First, transition layers and sparse textual anchors can be used as robustness diagnostics, especially for studying adversarial vulnerability and hallucination behavior.
Second, training strategies for concatenation models may benefit from earlier cross-modal feedback, anchor regularization, or layer-selective adaptation around the transition band.
Finally, layer-selective fine-tuning, adapters, or LoRA modules can be guided by the observed pathway differences instead of being placed uniformly across all layers.

% \section*{AI Use Statement}

% Generative AI tools were used to assist with manuscript editing, LaTeX formatting.
% We did not use generative AI tools to generate synthetic datasets, develop theoretical models or conceptual frameworks, formulate mathematical claims, provide critical ingredients for proofs, write proofs, propose or refine hypotheses, design the core experiments, clean or reformat datasets, or interpret experimental results.
% All AI-assisted text, figures, tables, code edits, citations, and formatting changes were reviewed by the authors, who take full responsibility for the final content of this work.

\section*{Acknowledgments}
This work was supported by a startup grant from the Mohamed bin Zayed University of Artificial Intelligence (MBZUAI) under grant number 11-101-5050-62410-CMS000-BF0111-KD234-00.

\bibliography{ref}
\bibliographystyle{unsrt}

\newpage
\appendix

\section{Technical Appendices and Supplementary Material}
\label{app:technical_appendix}

Appendix~\ref{app:multi_knn} reports the Mutual KNN analysis, which verifies that the global CKA trend is also reflected in local neighborhood structures.
Appendix~\ref{app:intra_modality_cka} provides the individual-model visual and textual CKA curves underlying the alignment-decoupling summary.
Appendix~\ref{app:attention_routing_entropy} presents the complete attention routing and cross-modal entropy results for all evaluated models.
Appendix~\ref{app:intrinsic_dimension} provides additional intrinsic-dimensionality curves to further support the observed feature-space expansion patterns.
Appendix~\ref{app:cross_model_visual_similarity} reports cross-model visual CKA heatmaps at intermediate layers, showing that deep-layer representation convergence develops progressively across depth rather than appearing only at the deepest displayed layer.
Appendix~\ref{app:bootstrap} reports the bootstrap robustness analysis.

\subsection{Mutual KNN}
\label{app:multi_knn}

The global CKA curves in the introduction reveal the overall layer-wise similarity trend between visual and textual representations.
To further verify that this trend is not only reflected by kernel-level similarity, we additionally evaluate the local neighborhood consistency of representations using a Mutual KNN analysis~\cite{groger2026revisiting}.
For each model and each layer, we compute the nearest-neighbor structure of visual and textual representations under neighborhood size $k=10$.
Given a representation $z_i^{(l)}$ at layer $l$, we obtain its $k$ nearest neighbors and compare neighborhood consistency across layers and modalities.
Higher Mutual KNN consistency indicates that the local semantic structure of representations is more stable or better aligned.

\begin{figure}[H]
    \centering
    \includegraphics[width=0.8\linewidth]{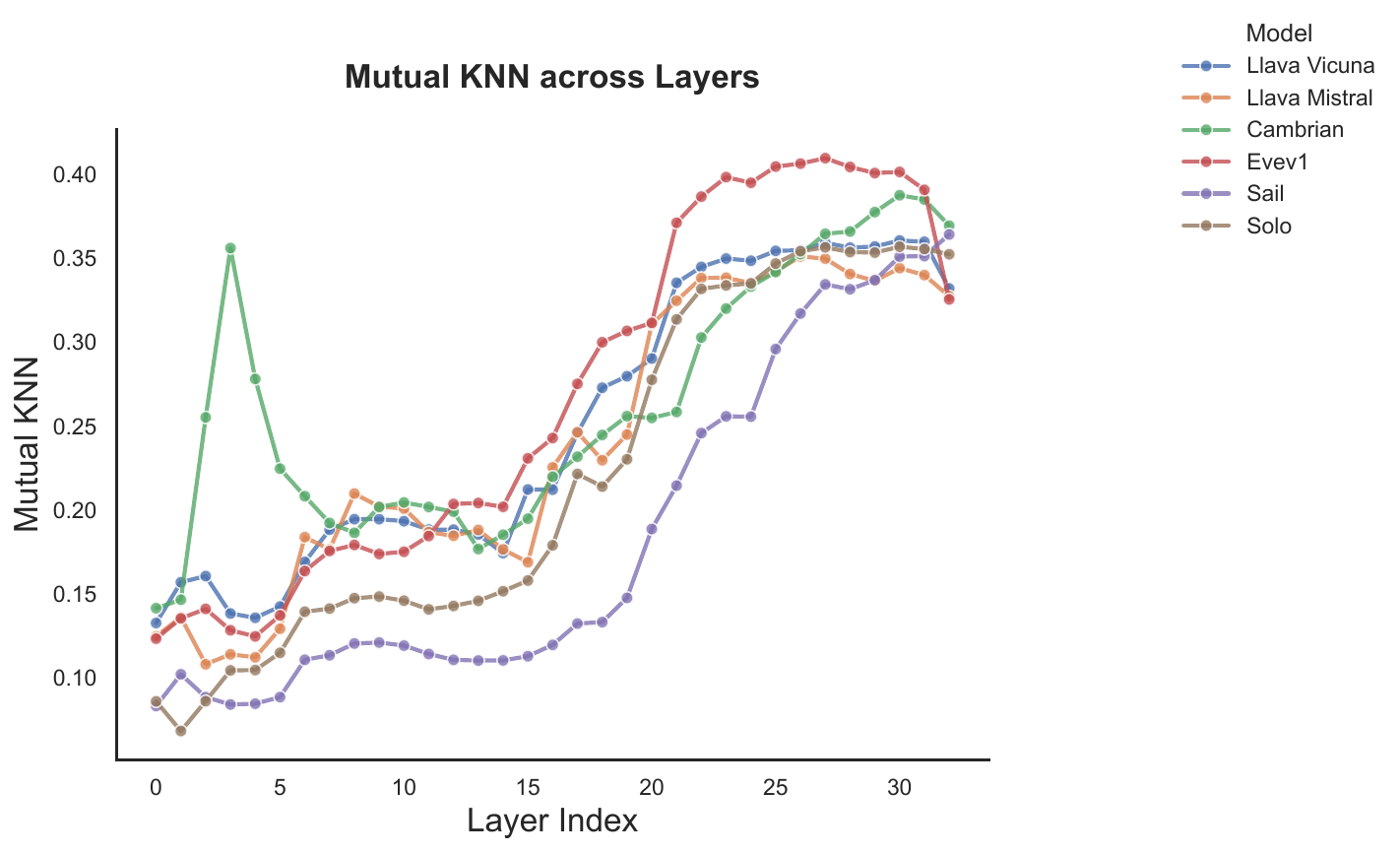}
    \caption{Supplementary Mutual KNN results. The observed trends are consistent with the global CKA analysis, suggesting that the layer-wise alignment pattern is also reflected in local neighborhood structure.}
    \label{fig:app_multi_knn}
\end{figure}

As shown in Figure~\ref{fig:app_multi_knn}, the Mutual KNN results generally follow the same trend as the global CKA curves.
Concatenation models exhibit a more obvious transition in the middle layers, while native models show more continuous representation evolution.
This supports the reliability of the global CKA observation in the main text.

\subsection{Per-model Intra-modality CKA}
\label{app:intra_modality_cka}

\begin{figure}[H]
    \centering
    \begin{subfigure}{\linewidth}
        \centering
        \includegraphics[width=\linewidth]{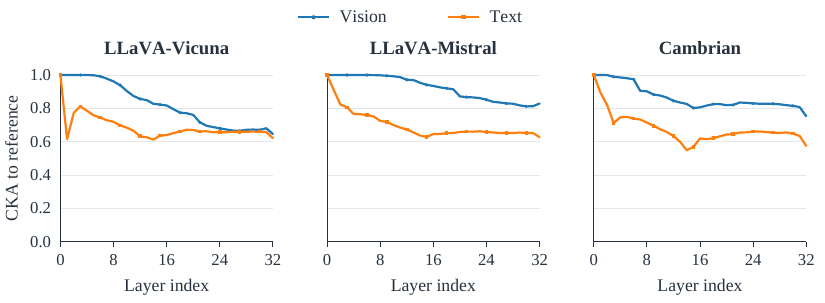}
        \caption{Concatenation models: LLaVA-Vicuna, LLaVA-Mistral, and Cambrian.}
    \end{subfigure}
    \par\medskip
    \begin{subfigure}{\linewidth}
        \centering
        \includegraphics[width=\linewidth]{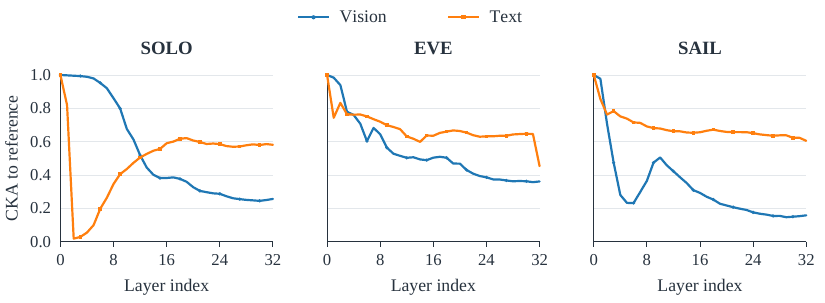}
        \caption{Native models: SOLO, EVE, and SAIL.}
    \end{subfigure}
    \caption{Per-model intra-modality CKA relative to the initial reference representation. Blue and orange curves denote visual and textual representations, respectively; both use solid lines. Layer indices are zero-based (0--32), consistent with Figure~\ref{fig:intra_modality_cka}.}
    \label{fig:app_intra_modality_cka}
\end{figure}

Figure~\ref{fig:app_intra_modality_cka} provides the detailed visual and textual trajectories underlying the family-level summary in Figure~\ref{fig:intra_modality_cka}.
The two panels separate concatenation and native models and retain each model's individual curves relative to the initial reference representation.
The detailed panels use the same zero-based layer indices (0--32) as Figure~\ref{fig:intra_modality_cka}, with the initial recorded representation at index 0 as the reference.

Visual representations remain closer to the reference in the shallow layers of the concatenation models, whereas the native models show stronger early visual changes.
Textual trajectories vary across models, including a pronounced early decrease followed by recovery in SOLO.
The individual curves therefore do not imply synchronized or monotonic changes in both modalities.

\subsection{Attention Routing and Cross-modal Entropy}
\label{app:attention_routing_entropy}

The main text compares attention distributions and directional median gaps across five models over layers 8--15, omitting SAIL because of its constrained attention path.
Here, we provide the corresponding layer-wise routing and normalized-entropy summaries for all six models, including SAIL.

\begin{figure}[t]
    \centering
    \begin{subfigure}[t]{0.48\linewidth}
        \centering
        \includegraphics[width=\linewidth]{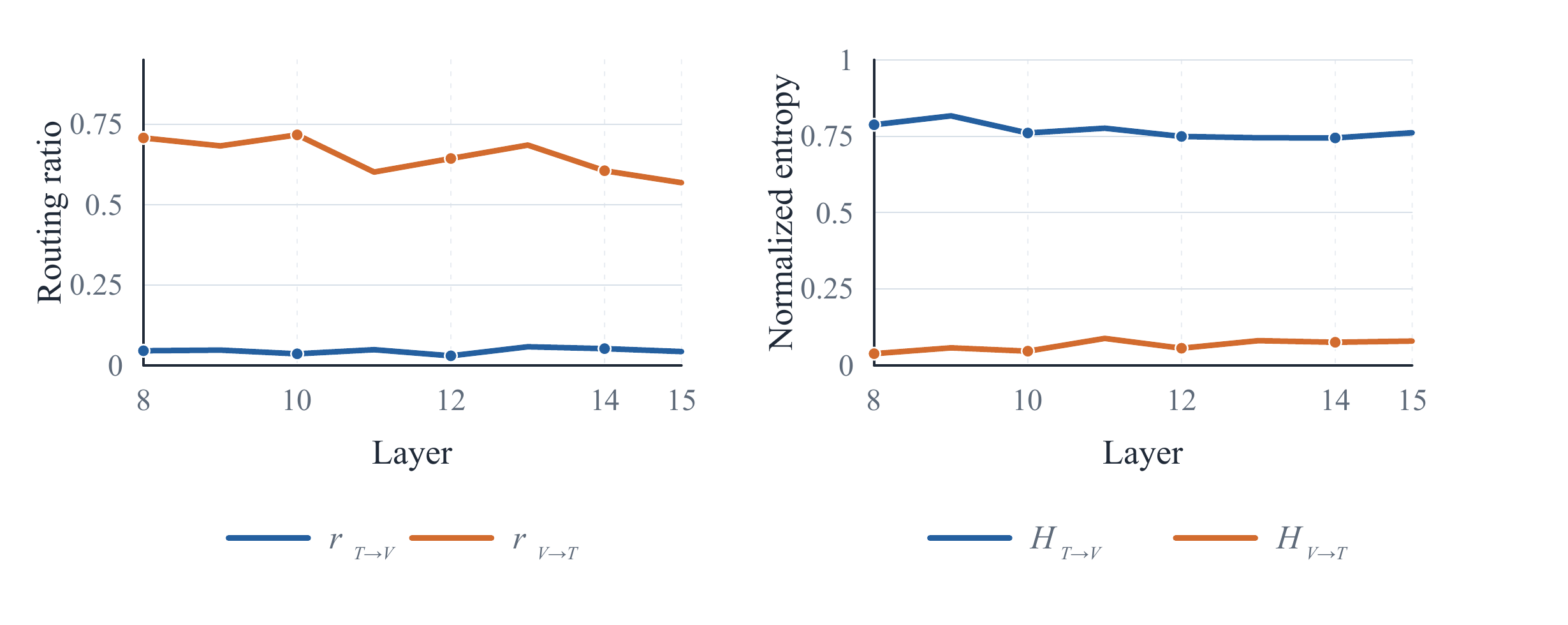}
        \caption{Cambrian.}
    \end{subfigure}
    \hfill
    \begin{subfigure}[t]{0.48\linewidth}
        \centering
        \includegraphics[width=\linewidth]{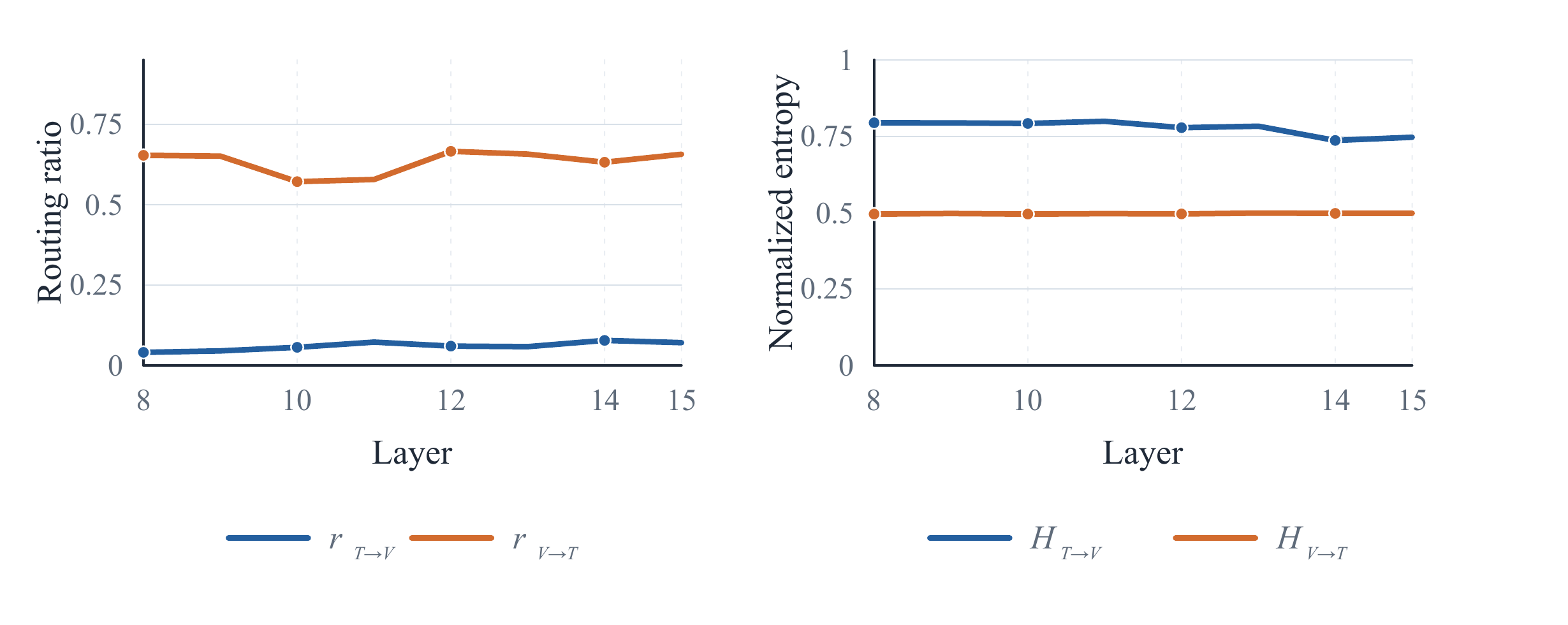}
        \caption{EVE.}
    \end{subfigure}
    
    \begin{subfigure}[t]{0.48\linewidth}
        \centering
        \includegraphics[width=\linewidth]{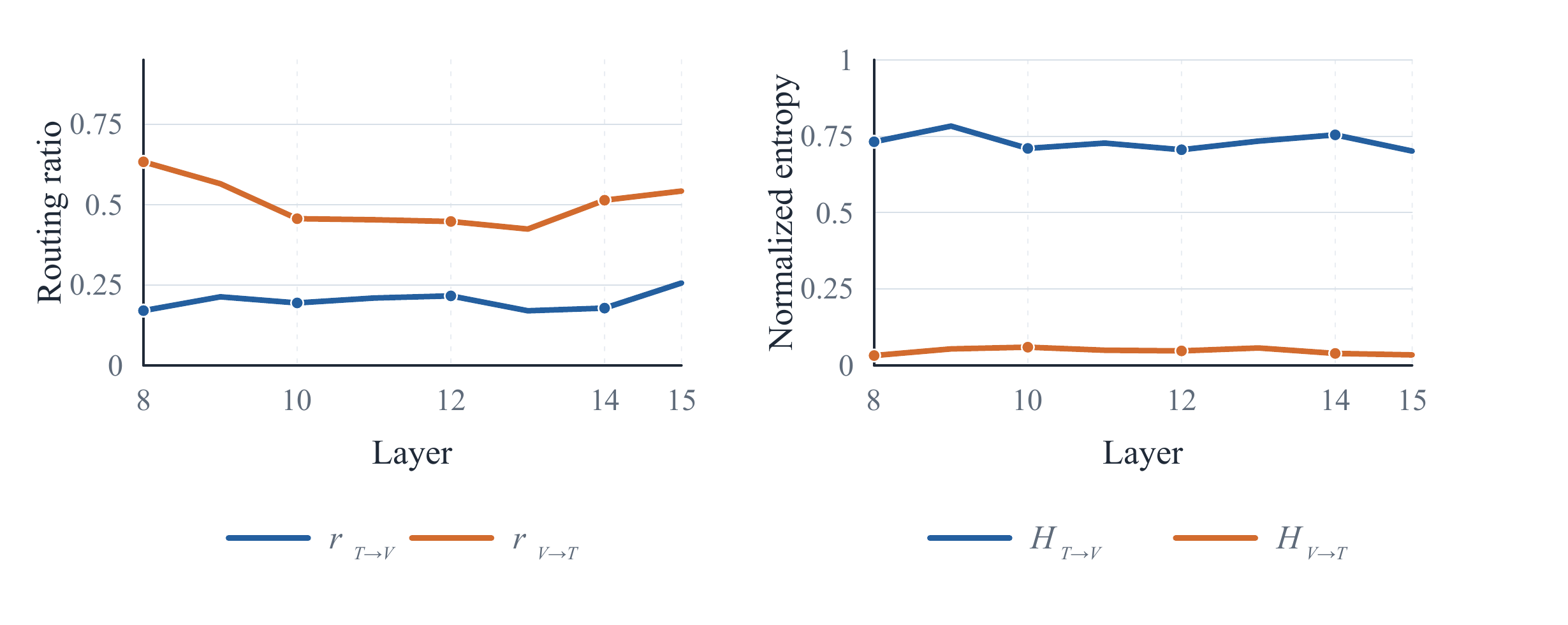}
        \caption{LLaVA-Mistral.}
    \end{subfigure}
    \hfill
    \begin{subfigure}[t]{0.48\linewidth}
        \centering
        \includegraphics[width=\linewidth]{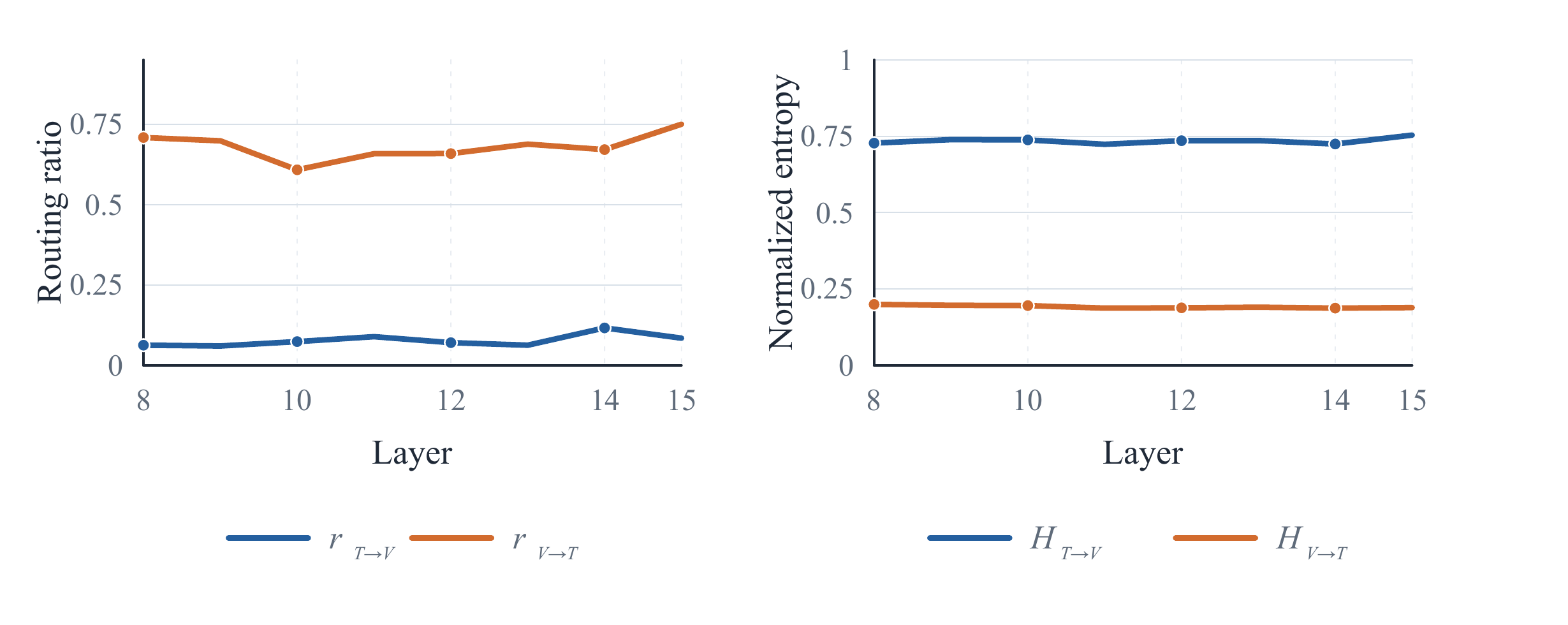}
        \caption{LLaVA-Vicuna.}
    \end{subfigure}
    
    \begin{subfigure}[t]{0.48\linewidth}
        \centering
        \includegraphics[width=\linewidth]{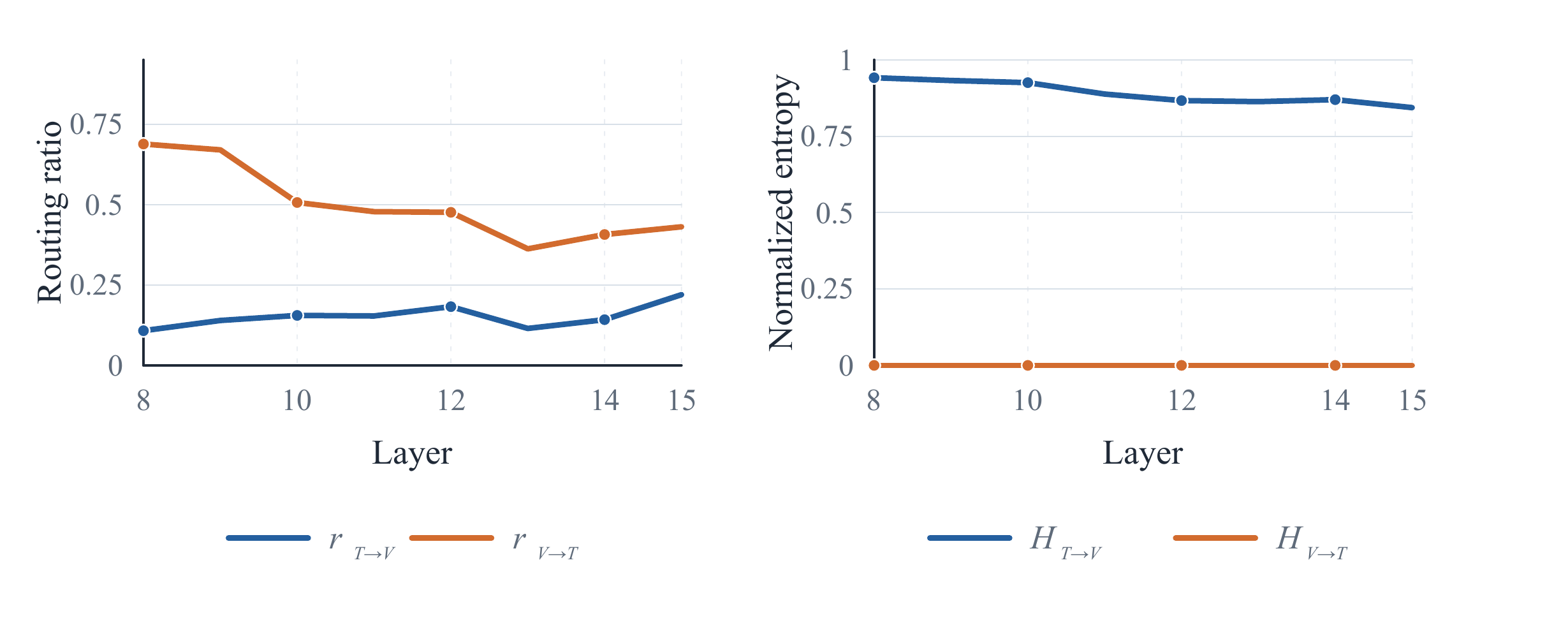}
        \caption{SAIL.}
    \end{subfigure}
    \hfill
    \begin{subfigure}[t]{0.48\linewidth}
        \centering
        \includegraphics[width=\linewidth]{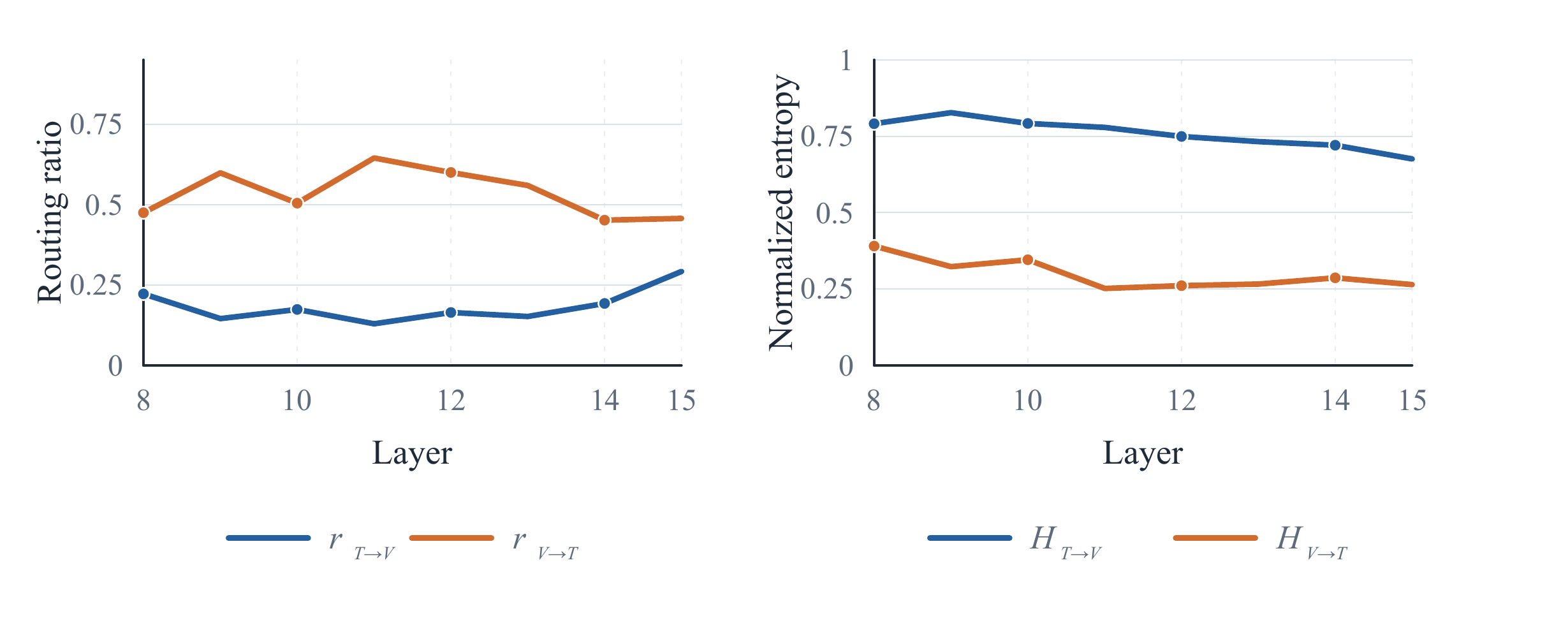}
        \caption{SOLO.}
    \end{subfigure}
    \caption{Attention routing ratios and per-sample normalized cross-modal entropy for all evaluated models over layers 8--15. SAIL has zero visual-to-text entropy because its hybrid attention path constrains that direction.}
    \label{fig:app_attention_routing_entropy}
\end{figure}

The layer-wise profiles in Figure~\ref{fig:app_attention_routing_entropy} complement the distribution summaries in Figure~\ref{fig:attention_routing_entropy}.
Concatenation models combine relatively high visual-to-text routing with low entropy in that direction across the analyzed range.
Native models exhibit heterogeneous attention profiles, including more diffuse visual-to-text attention in EVE and SOLO and a constrained visual-to-text entropy path in SAIL.
These curves characterize the sampled layer range; the onset of fusion is examined through the full-depth representation analyses.

\subsection{Intrinsic Dimension}
\label{app:intrinsic_dimension}

The intrinsic dimensionality analysis in the main text characterizes whether multimodal fusion corresponds to feature-space compression, expansion, or reorganization.
Here, we provide the original per-model visual and textual curves for additional models, followed by a threshold-sensitivity analysis for the two representative models used in the main paper.

\begin{figure}[t]
    \centering
    \begin{subfigure}{0.48\linewidth}
        \centering
        \includegraphics[width=\linewidth]{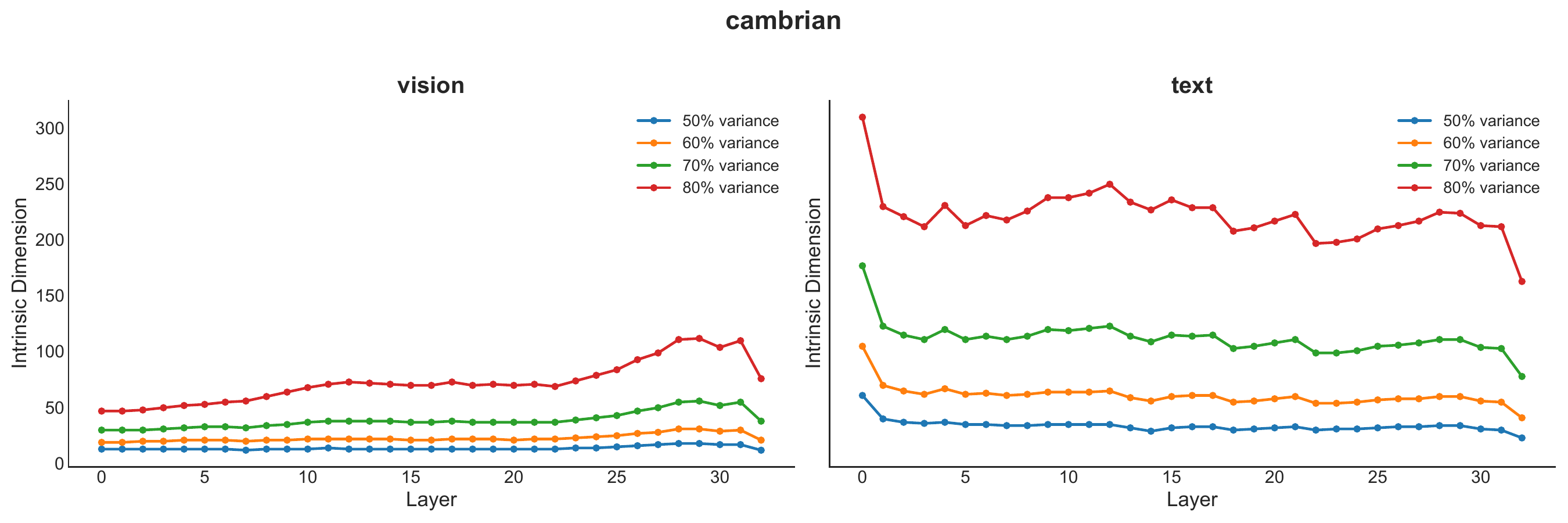}
        \caption{Cambrian.}
    \end{subfigure}
    \hfill
    \begin{subfigure}{0.48\linewidth}
        \centering
        \includegraphics[width=\linewidth]{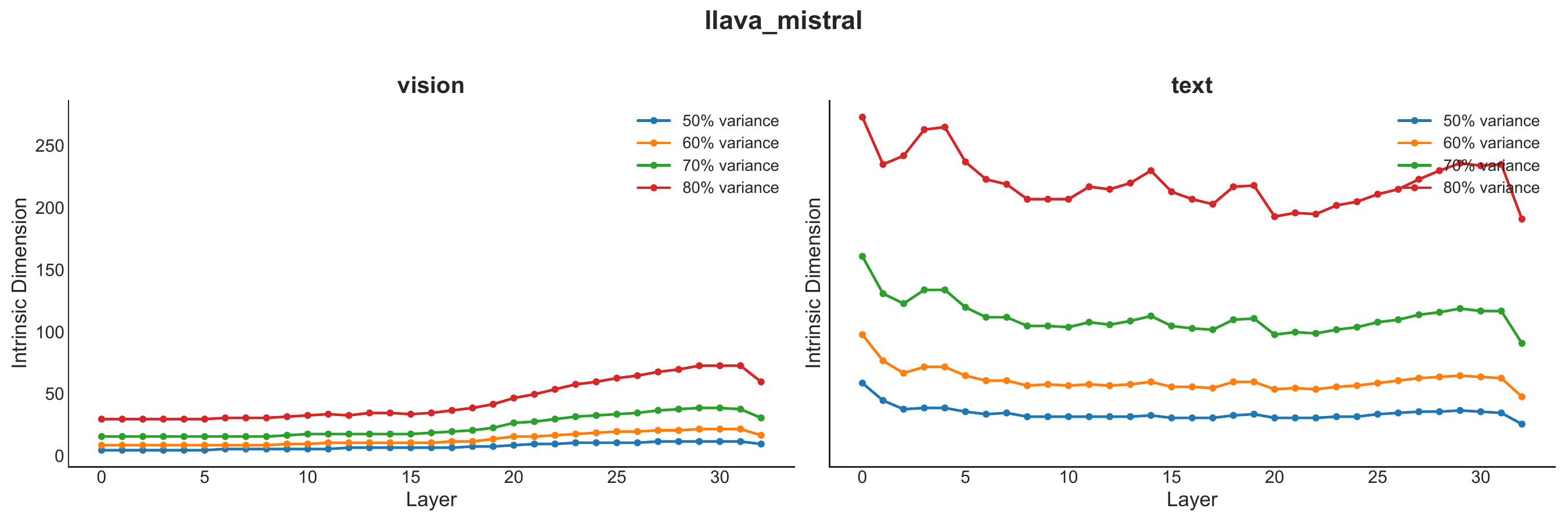}
        \caption{LLaVA-Mistral.}
    \end{subfigure}
    \hfill
    \begin{subfigure}{0.48\linewidth}
        \centering
        \includegraphics[width=\linewidth]{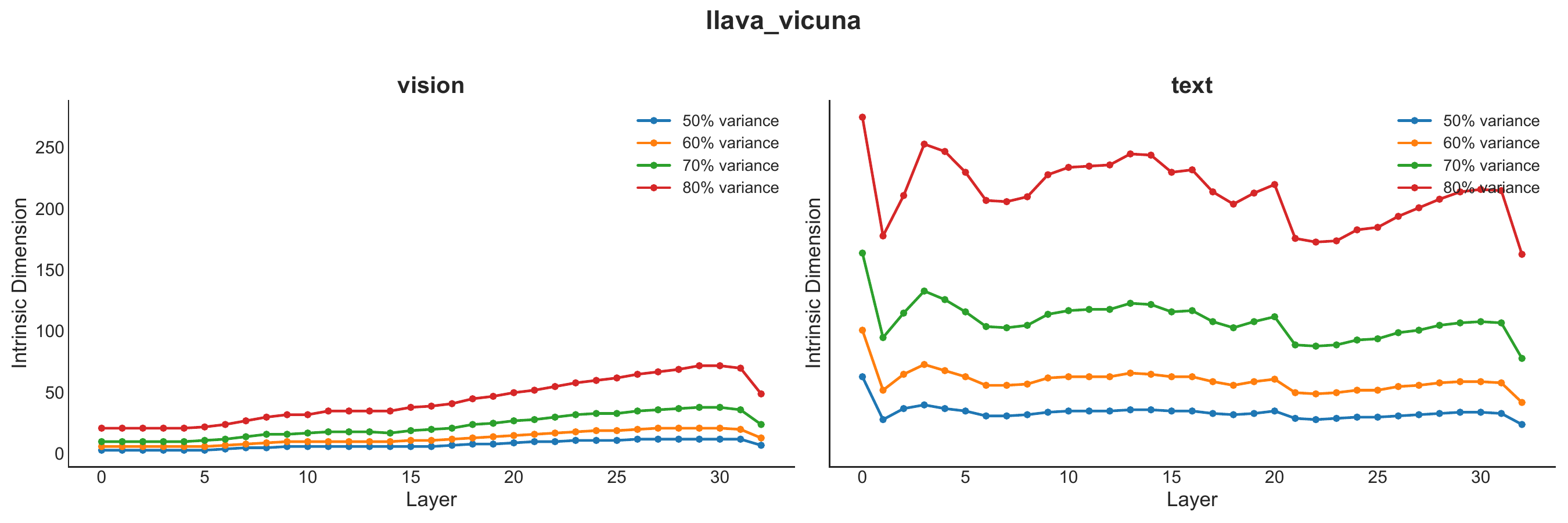}
        \caption{LLaVA-Vicuna.}
    \end{subfigure}
    \hfill
    \begin{subfigure}{0.48\linewidth}
        \centering
        \includegraphics[width=\linewidth]{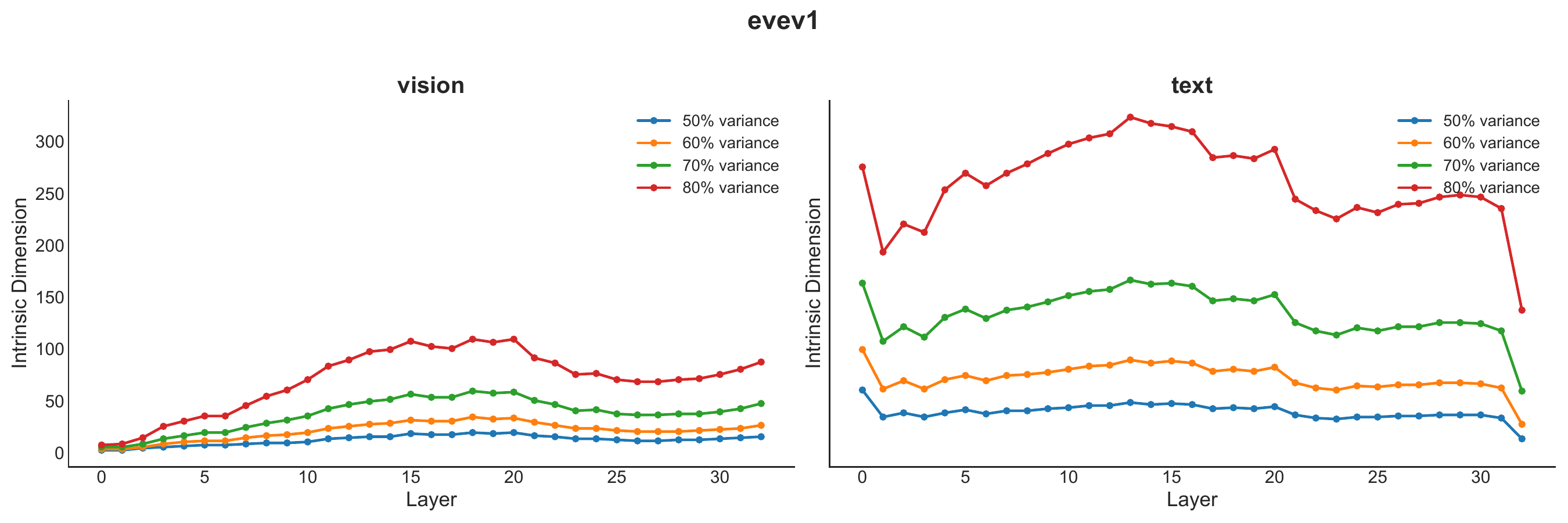}
        \caption{EVE.}
    \end{subfigure}
    \hfill
    \begin{subfigure}{0.48\linewidth}
        \centering
        \includegraphics[width=\linewidth]{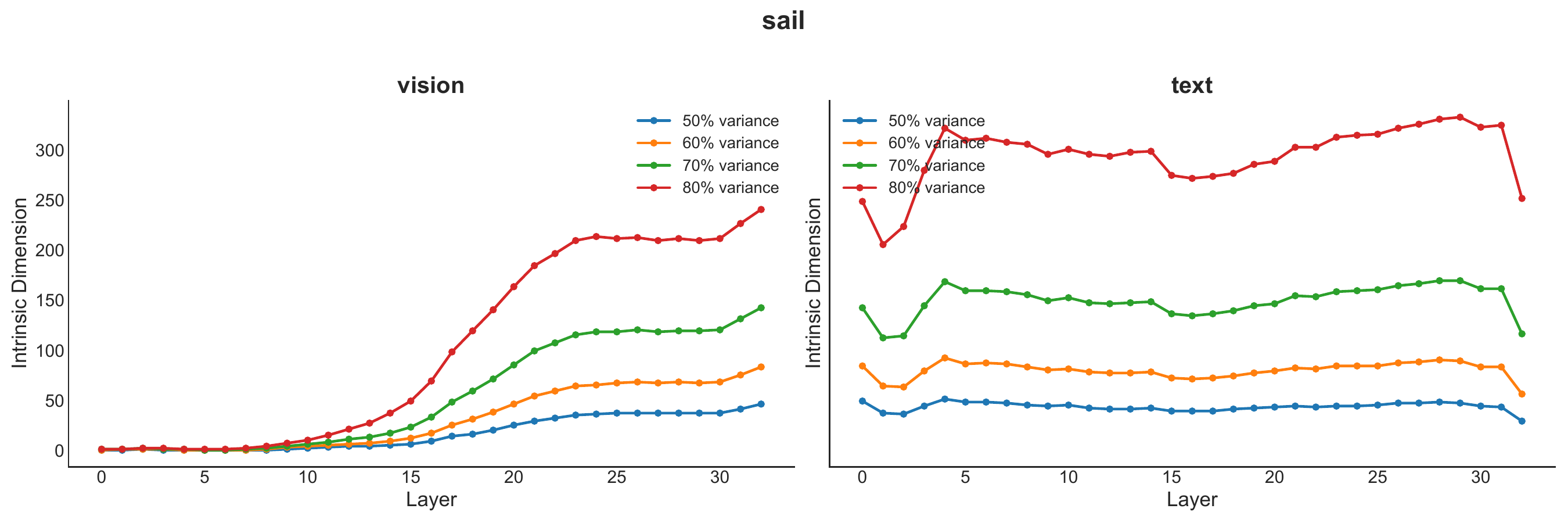}
        \caption{SAIL.}
    \end{subfigure}
    \hfill
    \begin{subfigure}{0.48\linewidth}
        \centering
        \includegraphics[width=\linewidth]{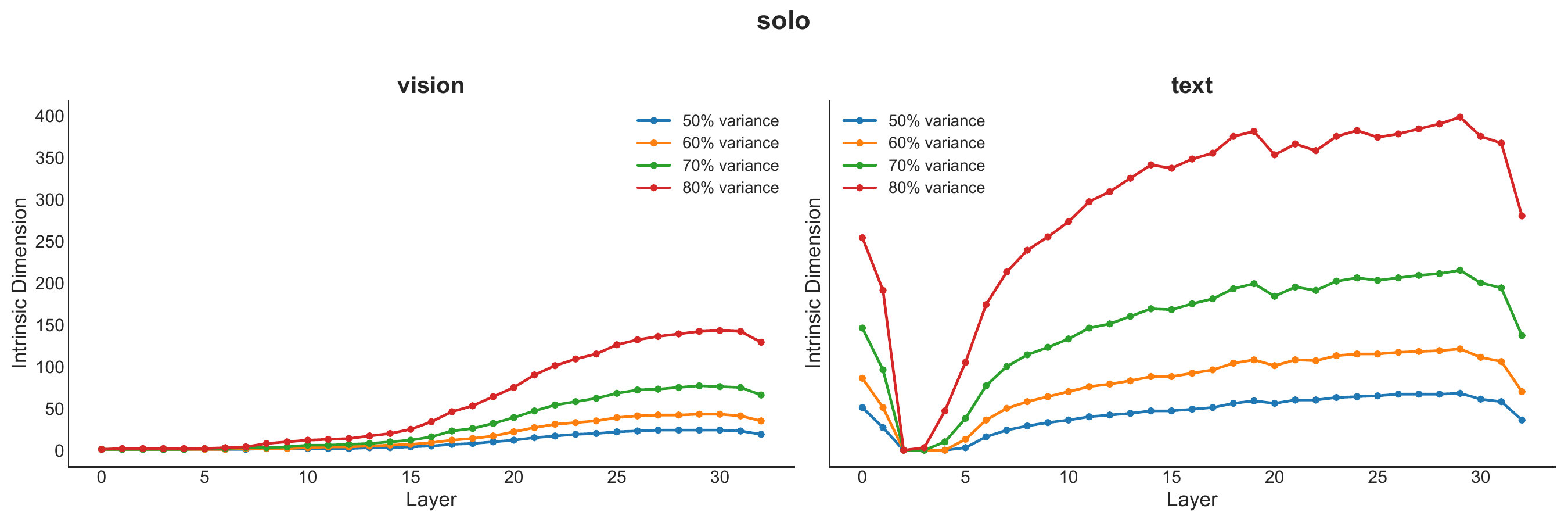}
        \caption{SOLO.}
    \end{subfigure}
    \caption{Supplementary intrinsic dimensionality results for additional models. Each panel uses the same layer-wise protocol as the main text and reports multiple variance preservation thresholds.}
    \label{fig:app_intrinsic_dimension}
\end{figure}

As shown in Figure~\ref{fig:app_intrinsic_dimension}, concatenation models generally show delayed visual dimensionality expansion, while native models show earlier and more continuous feature-space growth.
This is consistent with the main observation that concatenation models follow a text-first, vision-later fusion pathway, whereas native models perform earlier modality co-adaptation. 
Each model's plot contains multiple curves corresponding to different variance retention thresholds for the sensitivity analysis. The trends of all models remain consistent across varying thresholds, indicating that the observed patterns are not caused by a specific variance cutoff.

\subsection{Cross-model Visual Representation Similarity}
\label{app:cross_model_visual_similarity}

To further examine representation convergence across different MLLMs, we compute pairwise visual CKA similarity between models at shallow and deep layers.
This supplementary experiment is motivated by the Platonic Representation Hypothesis~\cite{huh2024platonic}: from the perspective of representation similarity, if different MLLMs gradually converge toward similar high-level visual semantics, their deep-layer visual representations should become more aligned.

\begin{figure}[t]
    \centering
    \begin{subfigure}[t]{0.48\linewidth}
        \centering
        \includegraphics[width=\linewidth]{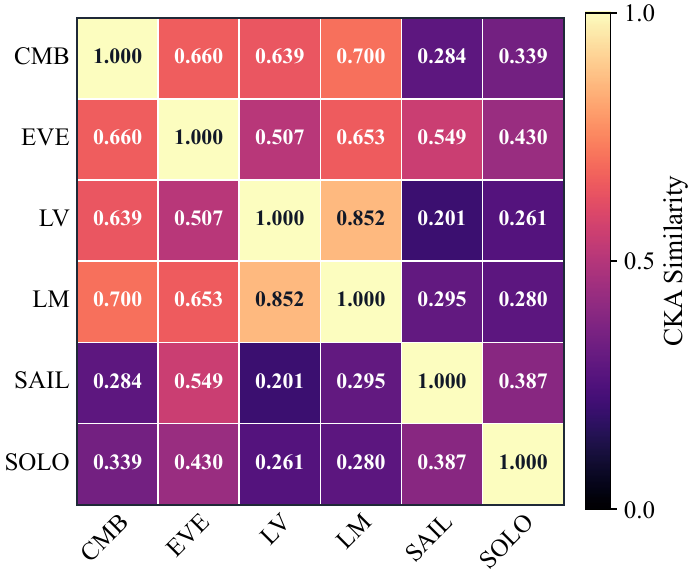}
        \caption{Layer 5.}
    \end{subfigure}
    \hfill
    \begin{subfigure}[t]{0.48\linewidth}
        \centering
        \includegraphics[width=\linewidth]{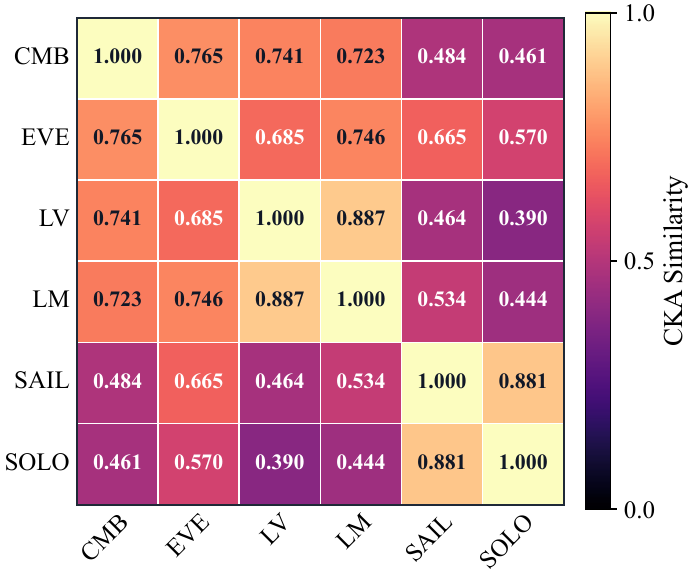}
        \caption{Layer 10.}
    \end{subfigure}
    
    \begin{subfigure}[t]{0.48\linewidth}
        \centering
        \includegraphics[width=\linewidth]{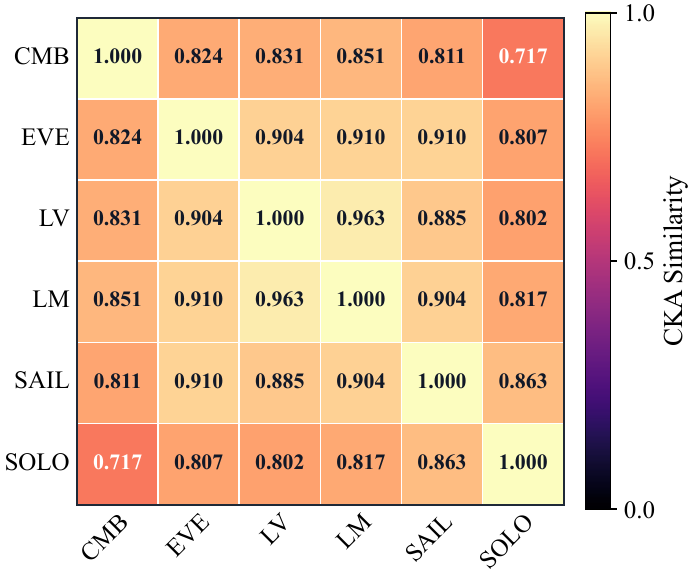}
        \caption{Layer 20.}
    \end{subfigure}
    \hfill
    \begin{subfigure}[t]{0.48\linewidth}
        \centering
        \includegraphics[width=\linewidth]{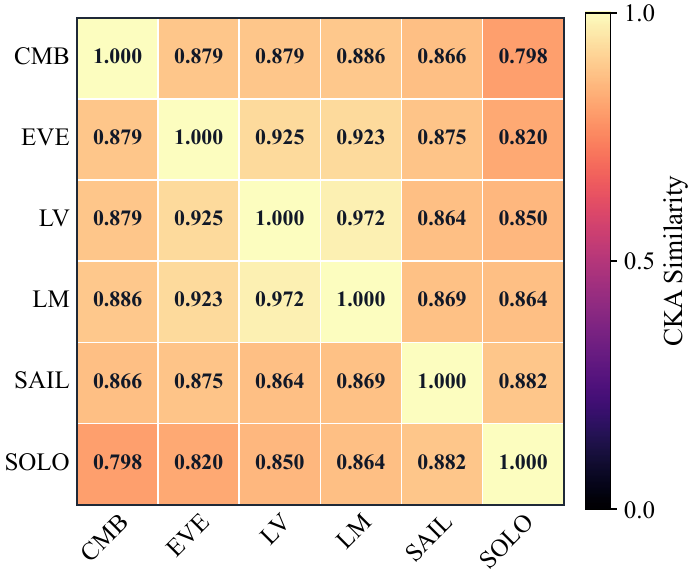}
        \caption{Layer 25.}
    \end{subfigure}
    \caption{Cross-model visual CKA heatmaps at selected appendix layers. A shared color scale and model order make the progressive increase in cross-model visual similarity easy to compare; lighter yellow cells indicate higher similarity.}
    \label{fig:app_cross_model_visual_cka}
\end{figure}

As shown in Figure~\ref{fig:app_cross_model_visual_cka}, visual representations are strongly architecture-dependent in shallow layers.
Models with similar visual processing strategies show higher similarity, while cross-paradigm similarities remain relatively low.
In contrast, deeper-layer heatmaps show substantially higher similarity across model pairs.
This result suggests that although different MLLMs start from different visual representation spaces, their deep visual representations become more aligned after multimodal fusion.
Therefore, this supplementary analysis provides representation-level evidence consistent with the Platonic Representation Hypothesis in the context of MLLMs.

\subsection{Bootstrap Robustness Analysis}
\label{app:bootstrap}

\begin{figure}[t]
    \centering
    \includegraphics[width=0.88\linewidth]{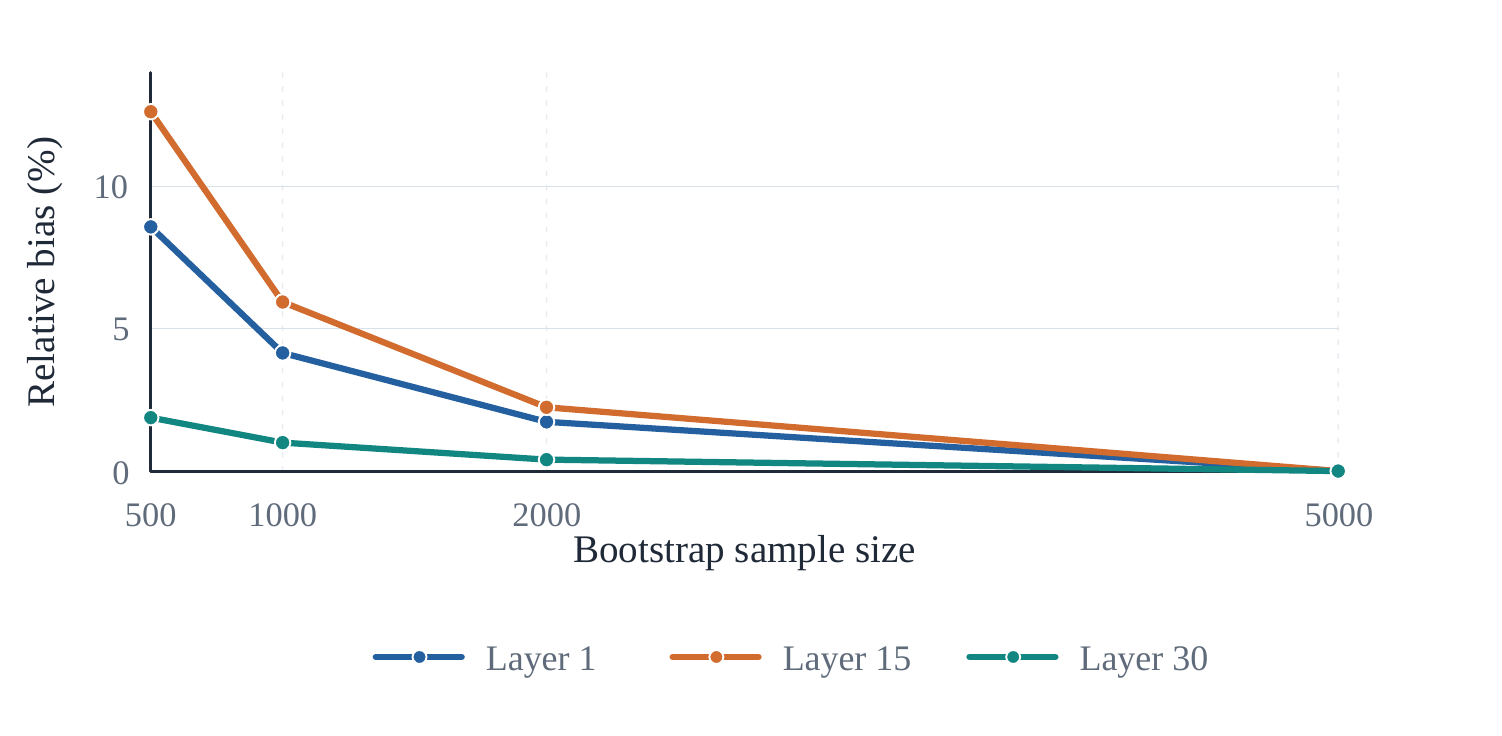}
    \caption{Bootstrap relative bias for Cambrian CKA at representative shallow, transition, and deep layers. The 5,000-sample estimate is the reference point for each layer.}
    \label{fig:bootstrap_bias}
\end{figure}

To quantify sampling uncertainty and evaluate robustness, we perform non-parametric bootstrap resampling.
We use a 5,000-pair analysis population, define four sample-size groups of 500, 1,000, 2,000, and 5,000 pairs, and conduct 1,000 independent random resamplings with replacement for each group, using image-text pairs as the resampling unit.
Layers 1, 15, and 30 are selected as representative layers, and the bootstrap mean under the 5,000-sample condition is used as the reference estimate for computing relative bias.
For layer $l$, relative bias (\%) is $100(\bar{c}_{n,l}-\bar{c}_{5000,l})/\bar{c}_{5000,l}$, where $\bar{c}_{n,l}$ is the bootstrap mean for sample size $n$.
It measures deviation from the 5,000-pair reference estimate, rather than bias relative to an unknown population CKA value.
All stochastic processes use a fixed random seed.
Figure~\ref{fig:bootstrap_bias} visualizes the relative-bias trend, while Table~\ref{tab:bootstrap_cambrian} reports the exact Cambrian CKA values.

\begin{table}[t]
\caption{Bootstrap robustness analysis for Cambrian in the Figure~\ref{fig:cka_comparison} experiment.}
\label{tab:bootstrap_cambrian}
\centering
\small
\begin{tabular}{lccc}
\toprule
Layer / sample size & Bootstrap mean & Standard error & Relative bias (\%) \\
\midrule
Layer 1 / 500 & 0.4679 & 0.0014 & 8.58 \\
Layer 1 / 1000 & 0.4488 & 0.0012 & 4.15 \\
Layer 1 / 2000 & 0.4384 & 0.0009 & 1.73 \\
Layer 1 / 5000 & 0.4309 & 0.0005 & 0.00 \\
Layer 15 / 500 & 0.3844 & 0.0013 & 12.63 \\
Layer 15 / 1000 & 0.3616 & 0.0011 & 5.94 \\
Layer 15 / 2000 & 0.3489 & 0.0008 & 2.24 \\
Layer 15 / 5000 & 0.3413 & 0.0004 & 0.00 \\
Layer 30 / 500 & 0.7540 & 0.0011 & 1.88 \\
Layer 30 / 1000 & 0.7475 & 0.0008 & 1.00 \\
Layer 30 / 2000 & 0.7431 & 0.0006 & 0.40 \\
Layer 30 / 5000 & 0.7401 & 0.0004 & 0.00 \\
\bottomrule
\end{tabular}
\end{table}

The standard errors remain small across all sample sizes, and the relative bias decreases as the sample size increases.
This indicates that the observed shallow-layer, transition-layer, and deep-layer CKA values are stable under bootstrap resampling.

\end{document}

%% file: math_commands.tex
\usepackage{amsmath,amsfonts,bm}

\def\eqref#1{equation~\ref{#1}}
\def\1{\bm{1}}

\DeclareMathAlphabet{\mathsfit}{\encodingdefault}{\sfdefault}{m}{sl}
\SetMathAlphabet{\mathsfit}{bold}{\encodingdefault}{\sfdefault}{bx}{n}

%% file: ref.bib
@inproceedings{kornblith2019similarity,
  title={Similarity of neural network representations revisited},
  author={Kornblith, Simon and Norouzi, Mohammad and Lee, Honglak and Hinton, Geoffrey},
  booktitle={International conference on machine learning},
  pages={3519--3529},
  year={2019},
  organization={PMLR}
}

@article{achiam2023gpt,
  title={Gpt-4 technical report},
  author={Achiam, Josh and Adler, Steven and Agarwal, Sandhini and Ahmad, Lama and Akkaya, Ilge and Aleman, Florencia Leoni and Almeida, Diogo and Altenschmidt, Janko and Altman, Sam and Anadkat, Shyamal and others},
  journal={arXiv preprint arXiv:2303.08774},
  year={2023}
}

@article{tu2024towards,
  title={Towards generalist biomedical AI},
  author={Tu, Tao and Azizi, Shekoofeh and Driess, Danny and Schaekermann, Mike and Amin, Mohamed and Chang, Pi-Chuan and Carroll, Andrew and Lau, Charles and Tanno, Ryutaro and Ktena, Ira and others},
  journal={Nejm Ai},
  volume={1},
  number={3},
  pages={AIoa2300138},
  year={2024},
  publisher={Massachusetts Medical Society}
}

@inproceedings{sima2024drivelm,
  title={Drivelm: Driving with graph visual question answering},
  author={Sima, Chonghao and Renz, Katrin and Chitta, Kashyap and Chen, Li and Zhang, Hanxue and Xie, Chengen and Bei{\ss}wenger, Jens and Luo, Ping and Geiger, Andreas and Li, Hongyang},
  booktitle={European conference on computer vision},
  pages={256--274},
  year={2024},
  organization={Springer}
}

@article{yin2024survey,
  title={A survey on multimodal large language models},
  author={Yin, Shukang and Fu, Chaoyou and Zhao, Sirui and Li, Ke and Sun, Xing and Xu, Tong and Chen, Enhong},
  journal={National Science Review},
  volume={11},
  number={12},
  pages={nwae403},
  year={2024},
  publisher={Oxford University Press}
}

@article{zhu2023minigpt,
  title={Minigpt-4: Enhancing vision-language understanding with advanced large language models},
  author={Zhu, Deyao and Chen, Jun and Shen, Xiaoqian and Li, Xiang and Elhoseiny, Mohamed},
  journal={arXiv preprint arXiv:2304.10592},
  year={2023}
}

@article{wang2024qwen2,
  title={Qwen2-vl: Enhancing vision-language model's perception of the world at any resolution},
  author={Wang, Peng and Bai, Shuai and Tan, Sinan and Wang, Shijie and Fan, Zhihao and Bai, Jinze and Chen, Keqin and Liu, Xuejing and Wang, Jialin and Ge, Wenbin and others},
  journal={arXiv preprint arXiv:2409.12191},
  year={2024}
}

@inproceedings{abnar2020quantifying,
  title={Quantifying Attention Flow in Transformers},
  author={Abnar, Samira and Zuidema, Willem},
  booktitle={Proceedings of the 58th Annual Meeting of the Association for Computational Linguistics},
  pages={4190--4197},
  year={2020},
  publisher={Association for Computational Linguistics},
  doi={10.18653/v1/2020.acl-main.385}
}

@inproceedings{clark2019does,
  title={What does BERT look at? an analysis of BERT’s attention},
  author={Clark, Kevin and Khandelwal, Urvashi and Levy, Omer and Manning, Christopher D},
  booktitle={Proceedings of the 2019 ACL workshop BlackboxNLP: analyzing and interpreting neural networks for NLP},
  pages={276--286},
  year={2019}
}

@misc{liu2024llavanext,
  title={Llavanext: Improved reasoning, ocr, and world knowledge},
  author={Liu, Haotian and Li, Chunyuan and Li, Yuheng and Li, Bo and Zhang, Yuanhan and Shen, Sheng and Lee, Yong Jae},
  year={2024}
}

@article{tong2024cambrian,
  title={Cambrian-1: A fully open, vision-centric exploration of multimodal llms},
  author={Tong, Shengbang and Brown, Ellis and Wu, Penghao and Woo, Sanghyun and Middepogu, Manoj and Akula, Sai C and Yang, Jihan and Yang, Shusheng and Iyer, Adithya and Pan, Xichen and others},
  journal={Advances in Neural Information Processing Systems},
  volume={37},
  pages={87310--87356},
  year={2024}
}

@inproceedings{radford2021learning,
  title={Learning transferable visual models from natural language supervision},
  author={Radford, Alec and Kim, Jong Wook and Hallacy, Chris and Ramesh, Aditya and Goh, Gabriel and Agarwal, Sandhini and Sastry, Girish and Askell, Amanda and Mishkin, Pamela and Clark, Jack and others},
  booktitle={International conference on machine learning},
  pages={8748--8763},
  year={2021},
  organization={PmLR}
}

@article{oquab2023dinov2,
  title={Dinov2: Learning robust visual features without supervision},
  author={Oquab, Maxime and Darcet, Timoth{\'e}e and Moutakanni, Th{\'e}o and Vo, Huy and Szafraniec, Marc and Khalidov, Vasil and Fernandez, Pierre and Haziza, Daniel and Massa, Francisco and El-Nouby, Alaaeldin and others},
  journal={arXiv preprint arXiv:2304.07193},
  year={2023}
}

@inproceedings{li2023blip,
  title={Blip-2: Bootstrapping language-image pre-training with frozen image encoders and large language models},
  author={Li, Junnan and Li, Dongxu and Savarese, Silvio and Hoi, Steven},
  booktitle={International conference on machine learning},
  pages={19730--19742},
  year={2023},
  organization={PMLR}
}

@article{chen2024solo,
  title={Solo: A single transformer for scalable vision-language modeling},
  author={Chen, Yangyi and Wang, Xingyao and Peng, Hao and Ji, Heng},
  journal={arXiv preprint arXiv:2407.06438},
  year={2024}
}

@article{diao2024unveiling,
  title={Unveiling encoder-free vision-language models},
  author={Diao, Haiwen and Cui, Yufeng and Li, Xiaotong and Wang, Yueze and Lu, Huchuan and Wang, Xinlong},
  journal={Advances in Neural Information Processing Systems},
  volume={37},
  pages={52545--52567},
  year={2024}
}

@inproceedings{lei2025scalability,
  title={The scalability of simplicity: Empirical analysis of vision-language learning with a single transformer},
  author={Lei, Weixian and Wang, Jiacong and Wang, Haochen and Li, Xiangtai and Liew, Jun Hao and Feng, Jiashi and Huang, Zilong},
  booktitle={Proceedings of the IEEE/CVF International Conference on Computer Vision},
  pages={20758--20769},
  year={2025}
}

@inproceedings{lin2014microsoft,
  title={Microsoft coco: Common objects in context},
  author={Lin, Tsung-Yi and Maire, Michael and Belongie, Serge and Hays, James and Perona, Pietro and Ramanan, Deva and Doll{\'a}r, Piotr and Zitnick, C Lawrence},
  booktitle={European conference on computer vision},
  pages={740--755},
  year={2014},
  organization={Springer}
}

@inproceedings{antol2015vqa,
  title={Vqa: Visual question answering},
  author={Antol, Stanislaw and Agrawal, Aishwarya and Lu, Jiasen and Mitchell, Margaret and Batra, Dhruv and Zitnick, C Lawrence and Parikh, Devi},
  booktitle={Proceedings of the IEEE international conference on computer vision},
  pages={2425--2433},
  year={2015}
}

@inproceedings{yue2024mmmu,
  title={Mmmu: A massive multi-discipline multimodal understanding and reasoning benchmark for expert agi},
  author={Yue, Xiang and Ni, Yuansheng and Zhang, Kai and Zheng, Tianyu and Liu, Ruoqi and Zhang, Ge and Stevens, Samuel and Jiang, Dongfu and Ren, Weiming and Sun, Yuxuan and others},
  booktitle={Proceedings of the IEEE/CVF conference on computer vision and pattern recognition},
  pages={9556--9567},
  year={2024}
}

@inproceedings{lee2025multimodal,
  title={How do multimodal foundation models encode text and speech? an analysis of cross-lingual and cross-modal representations},
  author={Lee, Hyunji and Liu, Danni and Sinhamahapatra, Supriti and Niehues, Jan},
  booktitle={Proceedings of the 2025 Conference of the Nations of the Americas Chapter of the Association for Computational Linguistics: Human Language Technologies (Volume 2: Short Papers)},
  pages={600--610},
  year={2025}
}

@article{liang2022mind,
  title={Mind the gap: Understanding the modality gap in multi-modal contrastive representation learning},
  author={Liang, Victor Weixin and Zhang, Yuhui and Kwon, Yongchan and Yeung, Serena and Zou, James Y},
  journal={Advances in Neural Information Processing Systems},
  volume={35},
  pages={17612--17625},
  year={2022}
}

@article{hendricks2021decoupling,
  title={Decoupling the role of data, attention, and losses in multimodal transformers},
  author={Hendricks, Lisa Anne and Mellor, John and Schneider, Rosalia and Alayrac, Jean-Baptiste and Nematzadeh, Aida},
  journal={Transactions of the Association for Computational Linguistics},
  volume={9},
  pages={570--585},
  year={2021},
  publisher={MIT Press One Rogers Street, Cambridge, MA 02142-1209, USA journals-info~…}
}

@inproceedings{zhang2025cross,
  title={Cross-modal information flow in multimodal large language models},
  author={Zhang, Zhi and Yadav, Srishti and Han, Fengze and Shutova, Ekaterina},
  booktitle={Proceedings of the Computer Vision and Pattern Recognition Conference},
  pages={19781--19791},
  year={2025}
}

@inproceedings{phang2021fine,
  title={Fine-tuned transformers show clusters of similar representations across layers},
  author={Phang, Jason and Liu, Haokun and Bowman, Samuel},
  booktitle={Proceedings of the Fourth BlackboxNLP Workshop on Analyzing and Interpreting Neural Networks for NLP},
  pages={529--538},
  year={2021}
}

@inproceedings{lu2025representation,
  title={Representation potentials of foundation models for multimodal alignment: A survey},
  author={Lu, Jianglin and Wang, Hailing and Xu, Yi and Wang, Yizhou and Yang, Kuo and Fu, Yun},
  booktitle={Proceedings of the 2025 Conference on Empirical Methods in Natural Language Processing},
  pages={16680--16695},
  year={2025}
}

@inproceedings{fang2026align,
  title={To Align or Not to Align: Strategic Multimodal Representation Alignment for Optimal Performance},
  author={Fang, Wanlong and Zhang, Tianle and Chan, Alvin},
  booktitle={Proceedings of the AAAI Conference on Artificial Intelligence},
  volume={40},
  pages={21056--21064},
  year={2026}
}

@article{li2021align,
  title={Align before fuse: Vision and language representation learning with momentum distillation},
  author={Li, Junnan and Selvaraju, Ramprasaath and Gotmare, Akhilesh and Joty, Shafiq and Xiong, Caiming and Hoi, Steven Chu Hong},
  journal={Advances in neural information processing systems},
  volume={34},
  pages={9694--9705},
  year={2021}
}

@article{almudevar2025aligning,
  title={Aligning multimodal representations through an information bottleneck},
  author={Almud{\'e}var, Antonio and Hern{\'a}ndez-Lobato, Jos{\'e} Miguel and Khurana, Sameer and Marxer, Ricard and Ortega, Alfonso},
  journal={arXiv preprint arXiv:2506.04870},
  year={2025}
}

@inproceedings{singh2022flava,
  title={Flava: A foundational language and vision alignment model},
  author={Singh, Amanpreet and Hu, Ronghang and Goswami, Vedanuj and Couairon, Guillaume and Galuba, Wojciech and Rohrbach, Marcus and Kiela, Douwe},
  booktitle={Proceedings of the IEEE/CVF conference on computer vision and pattern recognition},
  pages={15638--15650},
  year={2022}
}

@article{huh2024platonic,
  title={The platonic representation hypothesis},
  author={Huh, Minyoung and Cheung, Brian and Wang, Tongzhou and Isola, Phillip},
  journal={arXiv preprint arXiv:2405.07987},
  year={2024}
}

@article{groger2026revisiting,
  title={Revisiting the platonic representation hypothesis: An aristotelian view},
  author={Gr{\"o}ger, Fabian and Wen, Shuo and Brbi{\'c}, Maria},
  journal={arXiv preprint arXiv:2602.14486},
  year={2026}
}

@inproceedings{li2023evaluating,
  title={Evaluating object hallucination in large vision-language models},
  author={Li, Yifan and Du, Yifan and Zhou, Kun and Wang, Jinpeng and Zhao, Xin and Wen, Ji-Rong},
  booktitle={Proceedings of the 2023 conference on empirical methods in natural language processing},
  pages={292--305},
  year={2023}
}

@inproceedings{liu2024mmbench,
  title={Mmbench: Is your multi-modal model an all-around player?},
  author={Liu, Yuan and Duan, Haodong and Zhang, Yuanhan and Li, Bo and Zhang, Songyang and Zhao, Wangbo and Yuan, Yike and Wang, Jiaqi and He, Conghui and Liu, Ziwei and others},
  booktitle={European conference on computer vision},
  pages={216--233},
  year={2024},
  organization={Springer}
}

@inproceedings{duan2024vlmevalkit,
  title={Vlmevalkit: An open-source toolkit for evaluating large multi-modality models},
  author={Duan, Haodong and Yang, Junming and Qiao, Yuxuan and Fang, Xinyu and Chen, Lin and Liu, Yuan and Dong, Xiaoyi and Zang, Yuhang and Zhang, Pan and Wang, Jiaqi and others},
  booktitle={Proceedings of the 32nd ACM international conference on multimedia},
  pages={11198--11201},
  year={2024}
}
